\documentclass[preprint,12pt]{elsarticle}

\usepackage{amsmath,amssymb,amsfonts,mathtools}
\usepackage{amsthm}
\usepackage{booktabs,multirow,array}
\usepackage{graphicx}
\usepackage{algorithm}
\usepackage{algpseudocode}
\usepackage{enumitem}
\usepackage{xcolor}
\usepackage{microtype}
\usepackage{placeins}
\usepackage{capt-of}
\usepackage{etoolbox}
\usepackage{float}
\usepackage{url}
\usepackage{hyperref}
\hypersetup{
  hidelinks,
  pdftitle={Hybrid Federated Learning for Multimodal Fusion under Missing Modalities and Site Heterogeneity},
  pdfauthor={Xiaofei Wu; Rongmei Liang; Jian Qing Shi},
  pdfsubject={Research article},
  pdfkeywords={hybrid federated learning; information fusion; missing modalities; multiple-kernel learning; site heterogeneity}
}
\usepackage{cleveref}
\usepackage{tikz}
\usetikzlibrary{arrows.meta,positioning,fit,calc,backgrounds,decorations.pathreplacing}
\usepackage{pgfplots}
\pgfplotsset{compat=1.18}

\makeatletter
\def\ps@pprintTitle{\ps@plain}
\makeatother
\makeatletter
\patchcmd{\pprintMaketitle}{\par\vskip36pt}{\par\vskip20pt}{}{}
\patchcmd{\pprintMaketitle}{\Large\@title\par\vskip18pt}{\Large\@title\par\vskip14pt}{}{}
\makeatother
\journal{}

\newtheorem{assumption}{Assumption}
\newtheorem{theorem}{Theorem}
\newtheorem{proposition}{Proposition}
\newtheorem{corollary}{Corollary}

\theoremstyle{remark}

\crefname{assumption}{Assumption}{Assumptions}
\Crefname{assumption}{Assumption}{Assumptions}

\newcommand{\R}{\mathbb{R}}
\newcommand{\Hcal}{\mathcal{H}}
\newcommand{\Scal}{\mathcal{S}}
\newcommand{\Ical}{\mathcal{I}}
\newcommand{\Mcal}{\mathcal{M}}

\newcommand{\E}{\mathbb{E}}
\newcommand{\Prob}{\mathbb{P}}
\newcommand{\norm}[1]{\left\lVert #1\right\rVert}
\newcommand{\ip}[2]{\left\langle #1,#2\right\rangle}
\newcommand{\ind}{\mathbb{I}}
\newcommand{\softmaxrisk}{\mathsf{SR}_{\rho}}

\DeclareMathOperator*{\argmax}{arg\,max}

\begin{document}

\begin{frontmatter}

\title{Hybrid Federated Learning for Multimodal Fusion under Missing Modalities and Site Heterogeneity}

\author[ynu]{Xiaofei Wu}
\ead{xfwu1016@ynu.edu.cn}
\author[sustech]{Rongmei Liang}
\ead{liang_r_m@163.com}
\author[bnbu]{Jian Qing Shi\corref{cor1}}
\cortext[cor1]{Corresponding author.}
\ead{jianqingshi@bnbu.edu.cn}

\affiliation[ynu]{organization={Yunnan Key Laboratory of Statistical Modeling and Data Analysis, Yunnan University; School of Mathematics and Statistics, Yunnan University},
                addressline={Dongwaihuan South Road, Chenggong District},
                city={Kunming},
                state={Yunnan},
                postcode={650500},
                country={China}}
\affiliation[sustech]{organization={Department of Statistics and Data Science, Southern University of Science and Technology},
                     city={Shenzhen},
                     postcode={518055},
                     country={China}}
\affiliation[bnbu]{organization={Guangdong Provincial/Zhuhai Key Laboratory of Interdisciplinary Research and Application for Data Science, Beijing Normal-Hong Kong Baptist University},
                    addressline={2000 Jintong Road, Tangjiawan},
                    city={Zhuhai},
                    state={Guangdong},
                    postcode={519087},
                    country={China}}

\begin{abstract}
Multimodal fusion is studied under hybrid federation, where data are partitioned across samples and feature blocks. The proposed two-dimensional federated smooth multiple-kernel large-margin learning method (2D-FedSMKL) integrates availability-corrected modality scores, group-kernel regularization, and smooth worst-site risk aggregation. For fixed maps and propensities, the objective is convex and smooth. Curvature bounds and exact synchronous block-gradient equivalence are established for binary and multiclass scores, giving a conditional accelerated rate. An oracle-propensity binary risk bound and a propensity-perturbation result are also derived. Simulations and two public activity-recognition benchmarks assess site heterogeneity and modality absence. Worst-site aggregation improves worst-site performance under heterogeneity, including reversed and size-balanced designs. On the benchmarks, validation selects the correction; the selected model is not significantly worse than the strongest reference at the worst site, and group-regularization benefits vary by dataset.
\end{abstract}

% Highlights are supplied in highlights.txt as a mandatory separate upload.
% The optional graphical abstract is not embedded in the manuscript PDF.


\end{frontmatter}

\section{Introduction}
\label{sec:introduction}

\begin{figure}[H]
\centering
% Original native vector icons; no external artwork or bitmap inputs.
\begin{tikzpicture}[x=1cm,y=1cm,font=\small,line cap=round,line join=round,
 axis/.style={-{Latex[length=2.4mm]},thick,draw=black!75}]
% The two arrows encode ownership dimensions, not measured quantities.
\draw[axis] (3.3,2.4)--(11.7,2.4) node[midway,above=4pt,font=\small\bfseries] {Modality partition};
\draw[axis] (.3,-.2)--(.3,-4.9);
\node[rotate=90,font=\small\bfseries] at (0,-2.55) {Sample partition};
% Colored column frames identify four independent feature custodians.
\foreach \x/\name/\shade in {4/Baseline/blue,6.2/ECG/orange,8.4/Wrist/teal,10.6/Ankle/violet}{
 \fill[\shade!5] (\x-1,-4.8) rectangle (\x+1,-.2);
 \draw[\shade!65!black,dashed] (\x-1,-4.8) rectangle (\x+1,-.2);
 \node[font=\small\bfseries] at (\x,.22) {\name};
}
% Baseline record: clipboard, identity glyph and feature lines.
\begin{scope}[shift={(4,1.22)},draw=blue!65!black,line width=.75pt]
 \filldraw[fill=blue!6,rounded corners=2pt] (-.42,-.54) rectangle (.42,.5);
 \filldraw[fill=white,rounded corners=1pt] (-.18,.4) rectangle (.18,.6);
 \draw (-.19,.17) circle (.09);\draw (-.32,-.07) .. controls (-.32,.05) and (-.06,.05) .. (-.06,-.07);
 \draw (.03,.21)--(.28,.21) (.03,.03)--(.28,.03) (-.27,-.26)--(.27,-.26) (-.27,-.4)--(.18,-.4);
\end{scope}
% ECG monitor with screen, pulse trace, stand and electrode lead.
\begin{scope}[shift={(6.2,1.22)},draw=orange!80!black,line width=.75pt]
 \filldraw[fill=orange!6,rounded corners=2pt] (-.6,-.28) rectangle (.48,.53);
 \draw[rounded corners=1pt] (-.5,-.12) rectangle (.36,.4);
 \draw (-.44,.1)--(-.24,.1)--(-.14,.25)--(-.04,-.04)--(.08,.19)--(.2,.1)--(.31,.1);
 \draw (-.06,-.28)--(-.06,-.47) (-.32,-.47)--(.2,-.47);
 \draw (.48,.12).. controls (.83,.14) and (.82,-.39)..(.62,-.41);\draw (.58,-.41) circle (.055);
\end{scope}
% Wrist-mounted inertial sensor: strap, case and motion axes.
\begin{scope}[shift={(8.4,1.22)},draw=teal!75!black,line width=.75pt]
 \filldraw[fill=teal!7,rounded corners=1pt] (-.19,-.62) rectangle (.19,.62);
 \filldraw[fill=white,rounded corners=3pt] (-.42,-.32) rectangle (.42,.32);
 \filldraw[fill=teal!8,rounded corners=1pt] (-.3,-.2) rectangle (.3,.2);
 \draw[-{Latex[length=1mm]}] (-.13,-.1)--(-.13,.12);\draw[-{Latex[length=1mm]}] (-.13,-.1)--(.17,-.1);
 \draw (.1,.07)--(.2,.17);\fill (.34,0) circle (.025);
\end{scope}
% Ankle unit: original lower-leg contour, cuff and sensor pod.
\begin{scope}[shift={(10.6,1.22)},draw=violet!70!black,line width=.75pt]
 \draw (-.24,.63)--(-.24,-.22)..controls(-.25,-.43)and(-.39,-.5)..(-.42,-.54)--(.47,-.54)..controls(.58,-.52)and(.51,-.37)..(.36,-.36)--(.12,-.26)--(.12,.63);
 \filldraw[fill=violet!8] (-.32,-.07) rectangle (.2,.22);
 \filldraw[fill=white,rounded corners=1pt] (-.22,-.15) rectangle (.13,.29);
 \draw (-.15,.02)--(-.09,.13)--(-.02,-.02)--(.05,.1);
\end{scope}
% Site groups: native clinic buildings, two patient records per group.
\foreach \y/\site/\shade in {-.925/A/blue,-2.525/B/orange,-4.125/C/teal}{
 \fill[\shade!6,rounded corners=2pt] (.85,\y-.65) rectangle (2.8,\y+.65);
 \begin{scope}[shift={(1.42,\y+.08)},draw=\shade!65!black,line width=.65pt]
  \filldraw[fill=white] (-.28,-.35) rectangle (.28,.35);
  \draw (-.36,.35)--(0,.51)--(.36,.35);
  \draw (-.07,-.35)--(-.07,-.12)--(.07,-.12)--(.07,-.35);
  \foreach \x in {-.17,.11}{\foreach \z in {0,.17}{\draw (\x,\z) rectangle (\x+.07,\z+.07);}}
 \end{scope}
 \node[font=\scriptsize\bfseries] at (1.42,\y-.48) {Site \site};
}
\foreach \y/\id in {-.6/1,-1.25/2,-2.2/3,-2.85/4,-3.8/5,-4.45/6}{
 \begin{scope}[shift={(2.17,\y)},draw=black!65,line width=.6pt]
  \draw (0,.14) circle (.07);\draw (0,.07)--(0,-.12) (-.11,-.02)--(.11,-.02) (0,-.12)--(-.09,-.23) (0,-.12)--(.09,-.23);
 \end{scope}
 \node[font=\scriptsize] at (2.55,\y) {P\id};
 \foreach \x in {4,6.2,8.4,10.6}{\draw[black!16] (\x-1,\y-.27) rectangle (\x+1,\y+.27);}
}
% Availability pattern is unchanged from v22 (17 observed, 7 unavailable).
\foreach \x/\y in {4/-.6,4/-1.25,4/-2.2,4/-2.85,4/-3.8,4/-4.45,6.2/-.6,6.2/-1.25,6.2/-2.85,6.2/-4.45,8.4/-.6,8.4/-2.2,8.4/-3.8,10.6/-1.25,10.6/-2.2,10.6/-3.8,10.6/-4.45}{
 \node[text=teal!75!black] at (\x,\y) {$\checkmark$};
}
\foreach \x/\y in {6.2/-2.2,6.2/-3.8,8.4/-1.25,8.4/-2.85,8.4/-4.45,10.6/-.6,10.6/-2.85}{
 \fill[black!9] (\x-.96,\y-.23) rectangle (\x+.96,\y+.23);
 \node[text=black!65] at (\x,\y) {$\times$};
}
\node[font=\small] at (6.3,-5.4) {$\checkmark$ observed\qquad $\times$ missing};
\end{tikzpicture}
\caption{Illustrative sample and modality partitions in hybrid federated fusion. Sites group distinct patient records vertically; independently held modality blocks form the horizontal partition. Original device icons depict baseline records, ECG monitoring and wrist/ankle inertial sensors (IMUs). Check marks indicate observed blocks and crosses indicate record-level absence; the baseline remains observed. IDs, devices and availability patterns are schematic, not clinical study data. Computational row groups need not coincide with statistical sites. Raw feature blocks remain with their custodians. 2D-Fed denotes 2D-FedSMKL.}
\label{fig:architecture}
\end{figure}
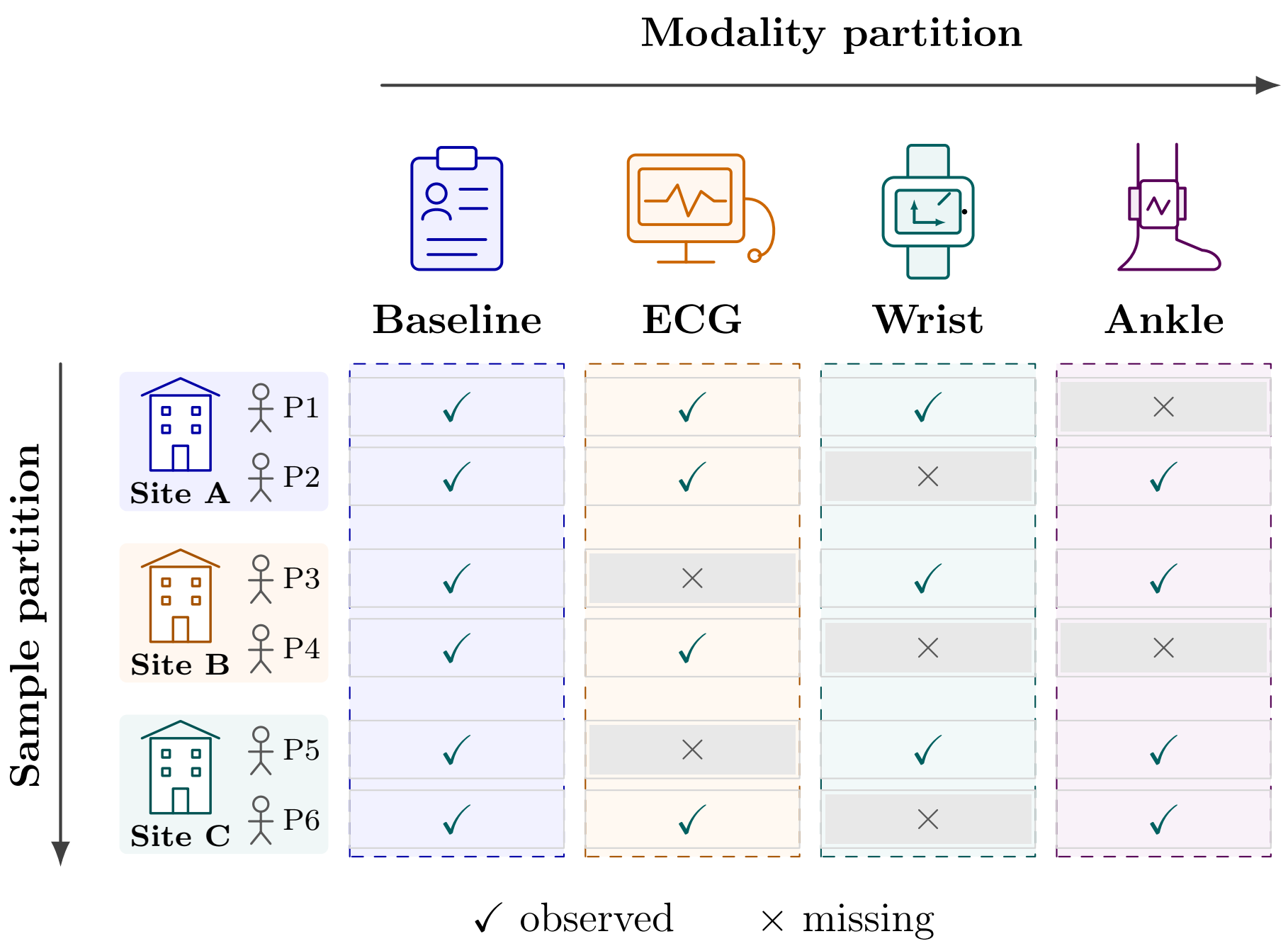

Multimodal information fusion combines sources such as wearable signals, physiological measurements, clinical examinations, and demographic variables. Their complementary information must be reconciled with differences in sampling, feature geometry, noise, and semantic content \cite{baltrusaitis2019multimodal,huang2024multimodal,sun2013survey,zhao2017multiviewoverview}. For example, free-living fatigue assessment has combined ECG, accelerometry, skin temperature, respiratory rate, and demographics \cite{bai2021fatigue}. These sources may be distributed and incomplete: multimodal medical studies can lack entire modalities \cite{wu2025omnifuse}, while federated systems partition data across samples, features, or both \cite{yang2019federated,kairouz2021advances,gao2023fedhd,zhang2024hybrid}. Figure~\ref{fig:architecture} illustrates the joint partition: sites contain different participant records, and separate custodians hold modality-specific feature blocks linked by aligned identifiers. Missing blocks are an additional record-level property, distinct from where features are held.

This layout creates four related difficulties. First, fusion must proceed without centralizing raw feature blocks, particularly in vertical or hybrid federation \cite{yang2019federated,kairouz2021advances,zhang2024hybrid}. Second, missingness often occurs by modality: an entire wearable recording, imaging study, or laboratory panel may be unavailable. Complete-case analysis then loses efficiency and can be biased when complete records are not an appropriate random subsample \cite{wu2025omnifuse,little2019statistical}. Third, hospitals can differ in patient mix, acquisition protocols, and endpoint prevalence. Such heterogeneity is a well-known challenge in federated optimization, and robust objectives can limit poor performance in shifted or underrepresented groups \cite{li2020federated,duchi2021learning,sagawa2020groupdro}. Finally, modalities may require distinct representations and similarity measures, motivating group-aware regularization beyond a shared penalty on concatenated features \cite{baltrusaitis2019multimodal,lanckriet2004learning}.

Activity recognition provides a concrete example of the decision task. In MHEALTH, a time window contains chest, ECG, wrist and ankle measurements, from which modality-specific summaries are extracted \cite{banos2014mhealthdata,banos2014mhealthdroid}. A support vector machine (SVM) learns a large-margin classification rule from these features and activity labels \cite{cortes1995support}. In the multiclass extension, one score is assigned to each activity and the largest score determines the prediction \cite{crammer2001multiclass}. Multiple-kernel learning (MKL) combines modality-specific score contributions and regularizes their relative influence \cite{lanckriet2004learning,rakotomamonjy2008simplemkl}. The fusion problem is to learn this rule when some window-level blocks are absent and different sites have different risks, while preserving the feature-custody constraints.

Our formulation combines four established components in a common objective. First, a softplus approximation to the unit-margin hinge loss permits smooth first-order optimization \cite{cortes1995support,lee2001ssvm,nesterov2005smooth}. Second, each observed modality score is weighted by an estimated observation propensity, allowing incomplete records to contribute to estimation \cite{little2019statistical,robins1994estimation}. Third, log-sum-exp aggregation smoothly approximates the maximum site risk and gives greater influence to sites with larger empirical risk \cite{duchi2021learning,sagawa2020groupdro,nesterov2005smooth}. Finally, a differentiable squared group-kernel penalty regularizes modalities and yields MKL weights through the connection between block norms and MKL perspective formulations \cite{rakotomamonjy2008simplemkl,bach2004multiple}. The feature maps are fixed throughout: the federated protocol covers the decision layer and its optimization, not representation learning.

For this hybrid federated setting, we propose \emph{two-dimensional federated smooth multiple-kernel large-margin learning} (2D-FedSMKL). The method makes the following contributions.

\begin{enumerate}[leftmargin=*,label=(\roman*)]
    \item We formulate availability-corrected multimodal decision fusion for simultaneous sample and feature partitions. The correction has two instantiations, an inverse-probability score with a risk guarantee and a conditional imputation, and is selected on validation data. The resulting objective is convex and couples modality-level regularization with smooth worst-site risk under fixed representations, and it has an explicit interface for evaluation without centralizing feature blocks.
    \item We establish target-specific curvature bounds and exact synchronous block-gradient equivalence for binary and multiclass scores. Standard accelerated optimization then gives a conditional convergence guarantee. An oracle-propensity binary risk bound and a training-sample propensity perturbation result quantify costs of site size and modality availability for fixed representations.
    \item We evaluate the framework through controlled simulations and two public activity-recognition benchmarks. The experiments examine predictive performance under site heterogeneity and modality absence, assess the contribution of the main modeling components, select the availability correction on validation data, and test the simulation mechanism under reversed and size-balanced site designs. We report where the framework matches or improves on the mean-risk and horizontal references, and where it does not.
\end{enumerate}

\Cref{sec:related} reviews missing-modality fusion, MKL, hybrid federated learning, and robust multicenter objectives. \Cref{sec:model} defines the score and objective, \Cref{sec:algorithm} develops the two-dimensional algorithm, and \Cref{sec:theory} gives the theoretical results. \Cref{sec:simulations} and \Cref{sec:public} report the controlled simulations and the two public benchmarks, and \Cref{sec:discussion} sets out the implications, limitations, and conclusions.

\section{Related work}
\label{sec:related}

\subsection{Incomplete multimodal information fusion}

Information may be fused at the data, feature, score, or decision level. Multimodal and multi-view learning address related problems of representation, alignment, fusion, translation, co-learning, and the use of complementary feature sets \cite{baltrusaitis2019multimodal,sun2013survey,zhao2017multiviewoverview}. In medical studies, intermediate or late fusion is often preferable when modalities differ in sampling rate, dimension, and quality-control requirements. Recent incomplete multi-view clustering methods first complete samples or graphs and then learn a consensus representation \cite{cai2024projected,cai2025incomplete,liu2025graph}; medical fusion methods have also begun to address severe modality absence and low-quality observations \cite{wu2025omnifuse}. Although this literature establishes missing modalities as a central problem, it mainly considers centralized representation learning or unsupervised clustering. We instead study supervised large-margin decision fusion when sites differ and data are distributed along both sample and feature dimensions.

Complete-case analysis is inefficient and may be biased unless complete records constitute an appropriate random subsample \cite{little2019statistical}. Inverse-probability weighting (IPW) addresses selection by weighting observed contributions by the reciprocal of their estimated observation probability \cite{robins1994estimation}. We apply this principle to modality-level scores: each observed score is reweighted under a stated missing-at-random condition. \Cref{sec:model} also defines a conditional-imputation instantiation, and \Cref{sec:public} selects between the two on validation data.

\subsection{Multiple-kernel large-margin fusion}

SVMs balance empirical margin loss against an RKHS norm when learning a separating function \cite{cortes1995support,scholkopf2002learning}. Kernel-based multi-view extensions include multi-view Laplacian SVMs, which combine agreement between views with margin learning in RKHSs \cite{sun2011mvlapsvm}. MKL learns a combination of base kernels. Its semidefinite, conic, and block-norm formulations established the connection between kernel weighting and group regularization \cite{lanckriet2004learning,rakotomamonjy2008simplemkl,bach2004multiple}. Related discriminative approaches include multi-kernel maximum-entropy discrimination \cite{chao2016multikernel}, while kernel alignment provides another criterion for selecting or combining kernels \cite{cortes2012algorithms}. Distributed online MKL has also been developed for networked streaming learners \cite{hong2023distributed}; that setting, however, assumes horizontally distributed examples rather than aligned feature blocks with site-dependent modality absence.

Related functional-data work jointly considers registration and clustering of multidimensional trajectories \cite{zeng2019simultaneous}. Our development assumes fixed modality feature maps. The simulations use linear modality kernels; nonlinear kernels can be approximated locally with random Fourier or Nystr\"om features, without sharing raw columns \cite{rahimi2007random,williams2001nystrom}. Previous generalization analyses of multi-view classifiers have also used a PAC-Bayesian perspective \cite{sun2017pacbayes}. Here, the analysis instead controls an availability-corrected worst-site risk and explicitly displays its dependence on the smallest site and the minimum modality-observation propensity. The squared sum of smoothed group norms is a differentiable analogue of the block norm obtained after optimizing the standard MKL perspective weights, so the fitted group norms can be interpreted directly as modality weights.

\subsection{Horizontal, vertical, and hybrid federated learning}

Horizontal federated learning aggregates models across disjoint samples with a shared feature space \cite{kairouz2021advances,mcmahan2017communication}. FedAvg and FedProx are prominent examples, with the latter designed for systems and statistical heterogeneity \cite{li2020federated}. Vertical federated learning instead aligns shared entities across parties that hold different features \cite{yang2019federated}.

Hybrid federated learning permits variation in both samples and features. FedHD addresses hybrid data through local updates and gradient tracking \cite{gao2023fedhd}; Zhang et al.\ use model matching to combine feature extractors and classifiers under sample and feature heterogeneity \cite{zhang2024hybrid}. Both works address sample and feature heterogeneity, as does the present formulation. The present target is a fixed-representation, availability-corrected worst-site decision objective evaluated through exact synchronous reductions. We assume aligned records and feature correspondence; arbitrary feature matching and entity alignment require a different formulation. Supplementary Material~S5 compares the assumptions, exchanged quantities, and guarantee objects.

Multimodal federated learning covers the wider design space surveyed by Huang et al.\ \cite{huang2024multimodal}, dual-clustered feature contrast under domain heterogeneity \cite{qiao2025fedccl}, and distillation-based embedding transfer \cite{le2025fedmekt}. Modality-completion and distillation methods address missingness at the representation level \cite{wu2025omnifuse,le2025fedmekt}; this paper fixes the representation and corrects observed scores instead. Our framework provides a convex decision objective with explicit assumptions, and the fitted group norms give the modality weights directly. Centralized and horizontal predictors are statistical references; the computational comparison uses ordinary and accelerated gradient methods on the same objective and row/column message interface, which isolates solver cost from the information available to each method.

\subsection{Site heterogeneity and robust objectives}

Multisource domain adaptation transfers information across labeled sources to an often-unlabeled target \cite{mansour2009domain,zhao2025multisource}. Multicenter training poses a related but different problem: every participating site is part of the target, so high pooled performance should not conceal failure at a smaller site. Distributionally robust optimization (DRO) formalizes robustness to changes in the data-generating distribution \cite{duchi2021learning,namkoong2017variance}, and group-DRO methods minimize or approximate the worst loss over prespecified groups \cite{sagawa2020groupdro}. We use an entropy-regularized maximization over site weights. Its closed form is the log-sum-exp of the empirical site risks, which smoothly interpolates between prevalence-weighted mean risk and worst-site risk and remains amenable to accelerated first-order optimization.

\section{Two-dimensional smooth large-margin fusion}
\label{sec:model}

\subsection{Data layout and notation}

Let $s\in\Scal=\{1,\ldots,S\}$ index statistical sites, with $n_s\geq1$ training observations at site $s$ and total $n=\sum_s n_s$. Let $m\in\Mcal=\{1,\ldots,M\}$ index modalities. For $i=1,\ldots,n_s$, observation $(s,i)$ has binary label $y_{si}\in\{-1,+1\}$ and potential measurement $x_{sim}\in\mathcal X_m$, where $\mathcal X_m$ is modality $m$'s measurement space. Write $x_{si}=(x_{si1},\ldots,x_{siM})$ and $r_{si}=(r_{si1},\ldots,r_{siM})$, with $r_{sim}\in\{0,1\}$ indicating whether the block is observed. An always-available baseline vector $z_{si}$, such as demographics or acquisition covariates, supports the observation propensity
\begin{equation}
    p_{sim}=\Prob(r_{sim}=1\mid z_{si},s).
    \label{eq:propensity}
\end{equation}
The fitted propensity is $\widehat p_{sim}$; $p_{\min}>0$ denotes the lower clipping threshold. Observation probabilities describe record-level availability, not worker connectivity. We apply inverse-probability weighting to modality-score contributions \cite{little2019statistical,robins1994estimation}.

For computation, reindex the pairs $(s,i)$ by global rows $1,\ldots,n$, and let $\mathcal N_s$ be the rows belonging to site $s$. In the finite-dimensional representation introduced below, modality $m$ has $d_m$ coordinates and $d=\sum_m d_m$. Let $X\in\R^{n\times d}$ denote the represented design matrix for block indexing; absent blocks have zero storage placeholders and retain their availability indicators. Row sets $\{\Ical_a\}_{a=1}^A$ partition $\{1,\ldots,n\}$, and column sets $\{\mathcal J_j\}_{j=1}^B$ partition $\{1,\ldots,d\}$. Worker $(a,j)$, $a=1,\ldots,A$, $j=1,\ldots,B$, holds $X[\Ical_a,\mathcal J_j]$ and the associated availability information. Statistical sites $S$ and computational row groups $A$ need not coincide. A modality may occupy one column group or span several. Feature-custodian records must be aligned before fitting \cite{yang2019federated}; entity matching is outside the present optimization problem.

\subsection{Availability-corrected additive kernel score}

\begin{figure}[t]
\centering
\begin{tikzpicture}[
    x=1cm,y=1cm,scale=1.4,transform shape,font=\small,
    mod/.style={draw=blue!45,rounded corners=2pt,fill=blue!6,minimum width=1.42cm,minimum height=0.62cm,align=center},
    missing/.style={draw=red!55,dashed,rounded corners=2pt,fill=red!3,minimum width=1.42cm,minimum height=0.62cm,align=center,text=red!65!black},
    wt/.style={draw=black!38,rounded corners=2pt,fill=orange!6,minimum width=1.42cm,minimum height=0.58cm,align=center},
    arr/.style={-{Latex[length=1.7mm]},semithick,draw=black!65}
]
\node[mod]     (m1) at (0,2.2) {M1 observed\\$r_{si1}=1$};
\node[missing] (m2) at (2.7,2.2) {M2 absent\\$r_{si2}=0$};
\node[mod]     (m3) at (5.4,2.2) {M3 observed\\$r_{si3}=1$};
\node[wt] (w1) at (0,0.8) {$a_{si1}=1/\widehat p_{si1}$};
\node[wt] (w2) at (2.7,0.8) {$a_{si2}=0$};
\node[wt] (w3) at (5.4,0.8) {$a_{si3}=1/\widehat p_{si3}$};
\draw[arr] (m1)--(w1);
\draw[arr] (m2)--(w2);
\draw[arr] (m3)--(w3);
\node[draw=green!45!black,rounded corners=3pt,fill=green!5,minimum width=7.5cm,minimum height=0.72cm,align=center] (sum) at (2.7,-0.8)
{$f_{si}=b+\sum_m a_{sim}\langle w_m,\phi_m(x_{sim})\rangle$};
\draw[arr] (w1)--(sum.north west);
\draw[arr] (w2)--(sum.north);
\draw[arr] (w3)--(sum.north east);
\end{tikzpicture}
\caption{Availability-corrected additive fusion for one sample. Observed modalities are inverse-probability weighted, whereas an absent block contributes zero rather than an imputed signal. Under the stated missing-at-random assumption and with the true observation propensities, the correction recovers each full-data linear score contribution in expectation. Estimated propensities are clipped to limit instability, which can introduce bias.}
\label{fig:ipw_score}
\end{figure}
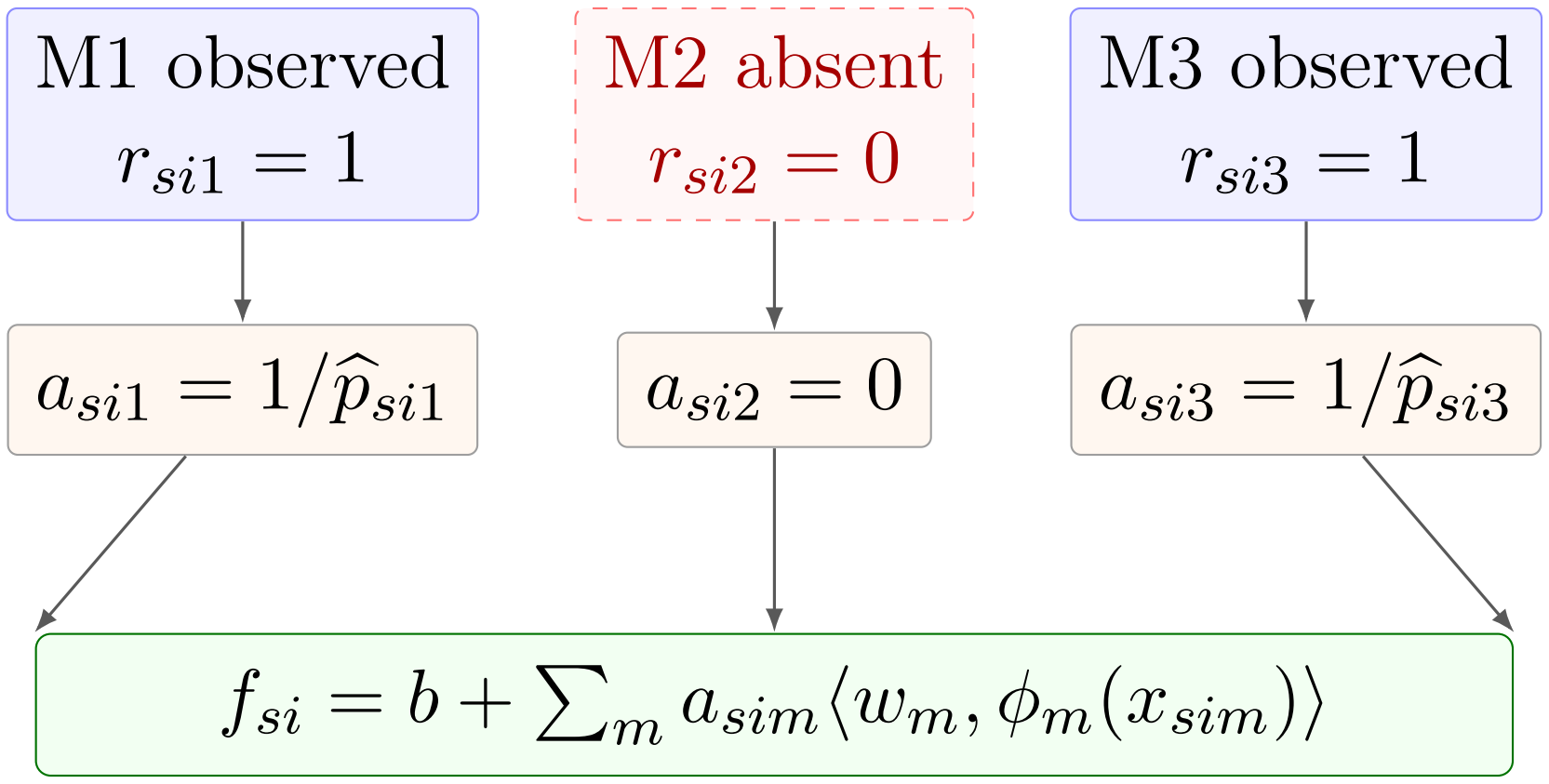

For modality $m$, let $\Hcal_m$ be a reproducing kernel Hilbert space (RKHS), with inner product $\langle\cdot,\cdot\rangle_{\Hcal_m}$ and norm $\|\cdot\|_{\Hcal_m}$. Its kernel and feature map satisfy \cite{scholkopf2002learning}
\[
\phi_m:\mathcal X_m\to\Hcal_m,\qquad
K_m(x,x')=\langle\phi_m(x),\phi_m(x')\rangle_{\Hcal_m}.
\]
The coefficient is $w_m\in\Hcal_m$, the intercept is $b\in\R$, and $w=(w_1,\ldots,w_M)$. With availability multiplier $a_{sim}$, the score is
\begin{equation*}
\begin{aligned}
    f_{w,b}(x_{si},r_{si})
    &=b+\sum_{m=1}^{M}a_{sim}
    \ip{w_m}{\phi_m(x_{sim})}_{\Hcal_m},\\
    a_{sim}&=\frac{r_{sim}}{\max(\widehat p_{sim},p_{\min})}.
\end{aligned}
\end{equation*}
Here $a_{sim}=0$ for an absent block, so no unavailable feature value is evaluated. For the oracle identity, replace the fitted denominator by the true propensity. For any fixed coefficient block $w_m$ and positive true propensity $p_{sim}>0$, the modality-level missing-at-random (MAR) condition $\Prob(r_{sim}=1\mid x_{sim},z_{si},s)=\Prob(r_{sim}=1\mid z_{si},s)=p_{sim}$ gives
\begin{equation}
    \E\!\left[\frac{r_{sim}}{p_{sim}}
    \ip{w_m}{\phi_m(x_{sim})}\,\middle|\,x_{sim},z_{si},s\right]
    =\ip{w_m}{\phi_m(x_{sim})}.
    \label{eq:ipw_unbiased}
\end{equation}
Equation~\eqref{eq:ipw_unbiased} concerns the linear score contribution under true propensities. A nonlinear loss applied to the sum of corrected scores generally changes this unbiasedness property. Accordingly, the objective below targets corrected-score risk. Fitted and clipped propensities can introduce bias, and small probabilities can amplify variability. If every modality is observed with probability one and the correction is set to one, the score reduces to ordinary additive MKL. Figure~\ref{fig:ipw_score} illustrates this correction without reconstructing an absent signal.

The correction is a module with two instantiations. The first is the inverse-probability score above. The second replaces an absent block with a conditional prediction from the always-observed block: for each optional block, a ridge regression is fitted on training rows where the block is observed and applied to the rows where it is absent, and the score is evaluated on the completed feature vector. Both instantiations use the same decision layer and hyperparameters, and the correction is selected on validation data by worst-site macro-F1, with ties resolved in favour of the inverse-probability form. Section~\ref{sec:theory} analyses the inverse-probability instantiation; the imputation instantiation targets the risk of the imputed score and is evaluated empirically in Section~\ref{sec:correction}.

For a finite-dimensional implementation, use $\phi_m(x)\in\R^{d_m}$ and $w_m\in\R^{d_m}$, stack $w\in\R^d$, and use Euclidean inner products and norms. The corrected feature vector is $\widetilde\phi_{si}=(a_{si1}\phi_1(x_{si1})^\top,\ldots,a_{siM}\phi_M(x_{siM})^\top,1)^\top$, with absent components zero. It includes the intercept coordinate. The analysis conditions on fixed feature and correction maps. Linear features can be used directly, whereas nonlinear kernels may be approximated locally with random Fourier or Nystr\"om features \cite{rahimi2007random,williams2001nystrom}. Define the smoothed group norm and the sum of these norms by
\begin{equation*}
    g_{m,\epsilon}(w_m)=\sqrt{\norm{w_m}_2^2+\epsilon^2},
    \qquad G_\epsilon(w)=\sum_{m=1}^{M}g_{m,\epsilon}(w_m),
\end{equation*}
with small $\epsilon>0$. 
\begin{samepage}
For fitted coefficient blocks $\widehat w$, the normalized modality weight is
\begin{equation}
    \widehat\beta_m=\frac{g_{m,\epsilon}(\widehat w_m)}
    {G_\epsilon(\widehat w)}.
    \label{eq:beta}
\end{equation}
\end{samepage}
For fixed $w$, minimizing $\sum_m(\|w_m\|_2^2+\epsilon^2)/\beta_m$ over positive kernel weights $\beta_m$ summing to one gives $G_\epsilon(w)^2$, with $\beta_m=g_{m,\epsilon}(w_m)/G_\epsilon(w)$, by Cauchy--Schwarz. As $\epsilon\downarrow0$, this recovers the standard block-norm/MKL perspective connection \cite{rakotomamonjy2008simplemkl,bach2004multiple}. Reported shares describe fitted coefficient blocks.

\subsection{Smooth unit-margin loss}

For a binary observation, write $t=y_{si}f_{w,b}(x_{si},r_{si})$ for its signed margin. Using $t$ as the scalar argument, the classical SVM uses the nondifferentiable hinge loss $(1-t)_+$ \cite{cortes1995support,scholkopf2002learning}. Smooth SVMs replace the positive-part operator with a differentiable approximation. The softplus below is the log-sum-exp approximation to $\max\{0,1-t\}$ and is closely related to the smoothing construction of Lee and Mangasarian \cite{lee2001ssvm,nesterov2005smooth}:
\begin{equation}
    \ell_\tau(t)=\frac{1}{\tau}\log\{1+\exp[\tau(1-t)]\},
    \qquad \tau>0.
    \label{eq:softplus_hinge}
\end{equation}
Its transition is centered on the SVM unit margin $t=1$, whereas the logistic-regression loss places its transition at $t=0$. Equivalently, $\ell_\tau(t)=\tau^{-1}\operatorname{LSE}\{0,\tau(1-t)\}$, where $\operatorname{LSE}$ denotes log-sum-exp. The standard smooth-maximum inequality therefore gives \cite{nesterov2005smooth}
\begin{equation*}
    0\leq \ell_\tau(t)-(1-t)_+\leq \frac{\log 2}{\tau},
\end{equation*}
and $\ell_\tau$ approaches hinge loss as $\tau\to\infty$. Its derivative magnitude is at most one, and its second derivative is at most $\tau/4$.

The empirical risk at site $s$ is
\begin{equation*}
    \widehat R_s(w,b)=\frac{1}{n_s}\sum_{i=1}^{n_s}
    \ell_\tau\!\left(y_{si}f_{w,b}(x_{si},r_{si})\right).
\end{equation*}

\subsection{Smooth worst-site aggregation}

\begin{figure}[t]
\centering
\begin{tikzpicture}[x=1cm,y=1cm,scale=1.4,transform shape,font=\small,
    site/.style={draw=blue!45,rounded corners=2pt,fill=blue!5,minimum width=1.25cm,minimum height=0.55cm,align=center},
    agg/.style={draw=orange!55!black,rounded corners=3pt,fill=orange!7,minimum width=2.75cm,minimum height=0.72cm,align=center},
    arr/.style={-{Latex[length=1.7mm]},semithick,draw=black!65}
]
\node[site] (s1) at (0,1.8) {Site 1\\$\widehat R_1$};
\node[site] (s2) at (2.05,1.8) {Site 2\\$\widehat R_2$};
\node[site] (s3) at (4.1,1.8) {Site 3\\$\widehat R_3$};
\node[agg] (a) at (2.05,0.42) {$\mathsf{SR}_{\rho}(\widehat R)$\\$q=\nabla_{\widehat R}\mathsf{SR}_{\rho}$};
\draw[arr] (s1)--node[left,font=\scriptsize] {$\pi_1$} (a.north west);
\draw[arr] (s2)--node[right,font=\scriptsize] {$\pi_2$} (a.north);
\draw[arr] (s3)--node[right,font=\scriptsize] {$\pi_3$} (a.north east);
\end{tikzpicture}
\caption{Smooth worst-site aggregation. Log-sum-exp provides a differentiable approximation to the largest site risk, and its gradient gives the adaptive weight $q_s$. Decreasing $\rho$ approaches worst-site training; increasing $\rho$ approaches $\pi$-weighted mean-risk training.}
\label{fig:site_risk}
\end{figure}
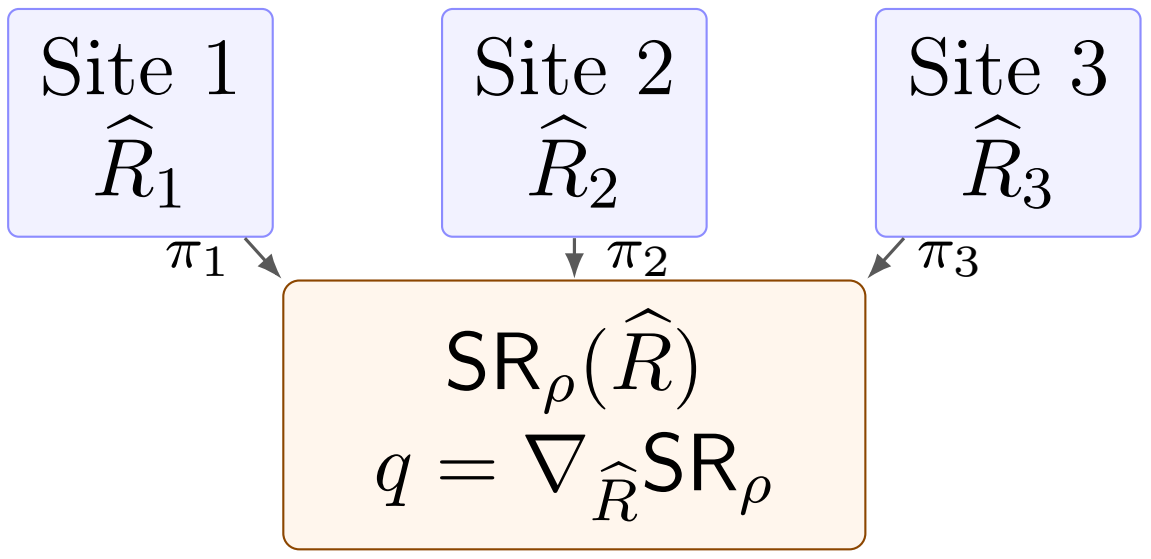

Average empirical risk can conceal poor performance in a small or shifted group; worst-group and worst-domain objectives are designed to address this problem \cite{duchi2021learning,sagawa2020groupdro}. Treating sites as prespecified groups, we replace the nonsmooth maximum site risk with a log-sum-exp approximation \cite{nesterov2005smooth}. Let $\pi_s>0$ and $\sum_s\pi_s=1$ denote reference site weights, ordinarily $n_s/n$, and define
\begin{equation*}
    \softmaxrisk(\widehat R_1,\ldots,\widehat R_S)
    =\rho\log\left\{\sum_{s=1}^{S}\pi_s
    \exp\left(\frac{\widehat R_s}{\rho}\right)\right\},
    \qquad \rho>0.
\end{equation*}
Its derivative with respect to $\widehat R_s$ is the adaptive site weight
\begin{equation}
    q_s(w,b)=\frac{\pi_s\exp\{\widehat R_s(w,b)/\rho\}}
    {\sum_{u=1}^{S}\pi_u\exp\{\widehat R_u(w,b)/\rho\}}.
    \label{eq:siteweight}
\end{equation}
The amplification factor $q_s/\pi_s$ increases with $\widehat R_s$; in particular, a site attaining the largest empirical risk has $q_s\geq\pi_s$. The absolute weights $q_s$ need not have the same ordering as the risks because they also depend on the reference weights $\pi_s$. This mechanism is the smooth analogue of upweighting groups with larger losses in group-DRO training \cite{sagawa2020groupdro} and is illustrated in \Cref{fig:site_risk}. It also satisfies
\begin{equation}
    \max_s\widehat R_s+\rho\log\pi_{\min}
    \leq \softmaxrisk(\widehat R)
    \leq \max_s\widehat R_s,
    \label{eq:softmax_bounds}
\end{equation}
where $\pi_{\min}=\min_s\pi_s$.
Thus \cref{eq:softmax_bounds} limits the smooth-maximum approximation gap by $\rho\log(1/\pi_{\min})$.

The proposed estimator minimizes
\begin{equation}
\begin{split}
    J(w,b)={}&\softmaxrisk\{\widehat R_1(w,b),\ldots,\widehat R_S(w,b)\}\\
    &+\frac{\lambda_0}{2}\left(\sum_{m=1}^{M}\norm{w_m}_2^2+\eta_b b^2\right)
    +\frac{\lambda_1}{2}G_\epsilon(w)^2.
\end{split}
\label{eq:objective}
\end{equation}
Here $\lambda_0,\lambda_1,\eta_b\geq0$ and $\tau,\rho,\epsilon>0$. The risk term controls the aggregation of site losses: as $\rho\to\infty$ it approaches the $\pi$-weighted mean, whereas $\rho\downarrow0$ emphasizes the largest site risk. This can protect a difficult site at the expense of average performance; smaller $\rho$ also increases curvature. The $\lambda_0$ term shrinks coefficients, with $\eta_b$ setting the relative intercept penalty. When both $\lambda_0$ and $\eta_b$ are positive, this penalty supplies strong convexity. Excessive shrinkage can increase bias. The $\lambda_1$ term couples modality norms and induces the fitted MKL shares. Its predictive effect depends on the data and missingness regime; Section~\ref{sec:public} reports a setting in which removing it improves performance.

The smoothing parameters have separate roles. Larger $\tau$ gives a sharper hinge approximation but increases loss curvature. Smaller $\epsilon$ more closely approximates the unsmoothed group norm, but increases the available curvature bound near zero blocks. These choices trade approximation accuracy against numerical conditioning and regularization. Section~\ref{sec:theory} states the resulting guarantees and the conditions under which they hold.

\subsection{Multiclass extension}

Crammer and Singer \cite{crammer2001multiclass} proposed a direct multiclass maximum-margin classifier. It assigns one score $f_c(x)$ to each of $C$ classes and predicts $\argmax_c f_c(x)$. Its per-example hinge loss can be written as
\begin{equation*}
    \ell^{\mathrm{CS}}(y,f)
    =\max\left\{0,\max_{c\neq y}\bigl(1+f_c-f_y\bigr)\right\}.
\end{equation*}
Applying the same log-sum-exp smoothing principle as in the binary case gives
\begin{equation}
    \ell_{\tau}^{\mathrm{mc}}(y,f)=
    \frac{1}{\tau}\log\left[1+\sum_{c\neq y}
    \exp\left\{\tau(1+f_c-f_y)\right\}\right],
    \qquad \tau>0.
    \label{eq:multiclass}
\end{equation}
The log-sum-exp approximation to a maximum satisfies \cite{nesterov2005smooth}
\begin{equation*}
    0\leq \ell_{\tau}^{\mathrm{mc}}(y,f)-\ell^{\mathrm{CS}}(y,f)
    \leq \frac{\log C}{\tau}.
\end{equation*}
Thus $\tau$ has the same interpretation as in \cref{eq:softplus_hinge}: larger values give a sharper approximation to the corresponding maximum-margin loss. For modality $m$, collect the class-specific coefficient vectors in $W_m=[w_{m1},\ldots,w_{mC}]$, use intercept $b\in\R^C$, and replace coefficient and intercept norms in \cref{eq:objective} by Frobenius and Euclidean norms, respectively. When $\lambda_0,\eta_b>0$, the quadratic penalty resolves the common-score-shift indeterminacy and ensures a unique minimizer. Partial class-score vectors remain additive across feature blocks. Supplementary Material~S4.7 proves convexity, strong convexity under these conditions, a multiclass curvature bound, and exact block-gradient equivalence. The population-risk results in \cref{sec:theory} concern binary classification; Section~\ref{sec:public} evaluates the multiclass extension empirically.

\section{Two-dimensional federated optimization}
\label{sec:algorithm}

\subsection{Score and gradient decomposition}

Write the finite-dimensional parameter as $\theta=(w^\top,b)^\top\in\R^{d+1}$, where $d=\sum_md_m$. We use the global row indexing of Section~\ref{sec:model}, suppressing the site index in $x_{im},r_{im},a_{im}$ and $y_i$. For sample $i\in\Ical_a$, column worker $j$ computes
\begin{equation*}
    z_{ij}=\sum_{m:\,\mathcal{D}_m\cap\mathcal{J}_j\neq\varnothing}
    a_{im}\ip{w_{m,j}}{\phi_{m,j}(x_{im})},
\end{equation*}
where $\mathcal D_m$ is the coordinate set of modality $m$, and $w_{m,j}$ and $\phi_{m,j}$ restrict its coefficients and features to $\mathcal D_m\cap\mathcal J_j$. Aggregation within the aligned row group gives
\begin{equation*}
    f_i=b+\sum_{j=1}^{B}z_{ij}.
\end{equation*}
The row coordinator evaluates
\begin{equation*}
    d_i=\frac{\partial\ell_\tau(y_if_i)}{\partial f_i}
    =-y_i\,\sigma\{\tau(1-y_if_i)\},
\end{equation*}
where $\sigma(u)=(1+e^{-u})^{-1}$. Let $\mathcal N_s$ denote the global row indices belonging to site $s$. After the empirical site risks are aggregated, the coordinator obtains $q_s$ from \cref{eq:siteweight}. The risk gradient for feature block $j$ is
\begin{equation*}
    \nabla_{w_j}\softmaxrisk
    =\sum_{s=1}^{S}\frac{q_s}{n_s}
    \sum_{i\in\mathcal N_s}d_i\,\widetilde\phi_j(x_i,r_i),
\end{equation*}
where $\widetilde\phi_j$ restricts the non-intercept coordinates of the corrected feature vector to $\mathcal J_j$. Each row worker forms the inner sum locally, after which only the resulting block gradient is reduced across row workers. The scalar risk gradient for the intercept is
\begin{equation*}
    g_b^{\rm risk}=\sum_{s=1}^{S}\frac{q_s}{n_s}\sum_{i\in\mathcal N_s}d_i,
\end{equation*}
and is reduced across row groups in the same synchronization stage.

Writing the two penalty terms in \cref{eq:objective} as $\operatorname{Reg}(w,b)$, their complete gradient contribution for modality $m$ and the intercept is
\begin{equation}
\begin{aligned}
    \nabla_{w_m}\operatorname{Reg}(w,b)
    &=\lambda_0w_m+\lambda_1G_\epsilon(w)
      \frac{w_m}{g_{m,\epsilon}(w_m)},\\
    \frac{\partial\operatorname{Reg}(w,b)}{\partial b}
    &=\lambda_0\eta_b b.
\end{aligned}
    \label{eq:reggradient}
\end{equation}
Computing $G_\epsilon$ requires one scalar norm per modality, or a reduction of squared norms if a modality spans several workers, followed by a scalar broadcast. Along with the additive reductions for scores and risk gradients, this operation keeps raw features local.

\subsection{Parallel accelerated updates}

We use accelerated gradient descent with a nondecreasing backtracking estimate \cite{beck2009fast}. Local computation alternates with structured aggregation, as in standard federated optimization \cite{kairouz2021advances,mcmahan2017communication}; unlike a horizontal scheme, however, the present layout partitions both samples and features \cite{gao2023fedhd,zhang2024hybrid}. For
\begin{equation*}
    Q_L(v,u)=J(u)+\ip{\nabla J(u)}{v-u}+\frac{L}{2}\norm{v-u}_2^2,
\end{equation*}
we accept a trial point $v$ when $J(v)\leq Q_L(v,u)$. Other solvers could use the same score-and-gradient interface, but the analysis below assumes synchronous accelerated updates so that distributed and centralized iterates can be compared exactly.

Each iteration has two parallel stages separated by reductions. During the \emph{parallel score stage}, every available worker $(a,j)$ evaluates partial scores for the samples in row block $a$. These scores are synchronized within the aligned row group because the complete score is needed to compute $f_i$, $d_i$, and the site losses. Once the site losses have been reduced and the adaptive weights $q_s$ are available, the \emph{parallel gradient stage} computes, for each row--column worker pair,
\begin{equation*}
    g_{aj}=\sum_{s=1}^{S}\frac{q_s}{n_s}
    \sum_{i\in\Ical_a\cap\mathcal N_s}d_i\,\widetilde\phi_j(x_i,r_i),
\end{equation*}

\noindent followed by the feature-block reduction $g_j=\sum_a g_{aj}$ and the scalar intercept reduction $g_b^{\rm risk}$. Computations for all row--column blocks are therefore concurrent between synchronization points.

\begin{algorithm}[H]
\caption{2D-FedSMKL with parallel, missing-modality-aware updates}
\label{alg:2dfed}
\footnotesize
\setlength{\baselineskip}{9.8pt}
\begin{algorithmic}[1]
\setlength{\itemsep}{0pt}
\setlength{\parskip}{0pt}
\Require Row partition $\{\Ical_a\}_{a=1}^{A}$; feature partition $\{\mathcal{J}_j\}_{j=1}^{B}$; propensity models.
\Require Parameters $\tau,\rho,\lambda_0,\lambda_1,\eta_b,\epsilon$ and backtracking factor $\gamma>1$.
\State Initialize $\theta_0=0$, extrapolated point $u_0=\theta_0$, $t_0=1$, and accepted estimate $\widehat L_{-1}>0$.
\For{$k=0,1,\ldots,T-1$}
    \Statex \textbf{Parallel score stage}
    \ForAll{row groups $a$ in parallel}
        \ForAll{column workers $j$ in parallel}
            \State Compute $z_{ij}(u_k)$ for $i\in\Ical_a$, observed blocks only. \Comment{partial score}
        \EndFor
        \State Synchronize column scores within row group $a$; compute $f_i$, $d_i$ and site-loss sums. \Comment{full score, $d_i$}
    \EndFor
    \State Synchronize site-loss scalars and compute $q_s$. \Comment{site weights}
    \Statex \textbf{Parallel gradient stage}
    \ForAll{feature blocks $j$ in parallel}
        \ForAll{row groups $a$ in parallel}
            \State Compute local risk-gradient block $g_{aj}$. \Comment{block gradient}
        \EndFor
        \State Reduce $g_j\gets\sum_{a=1}^{A}g_{aj}$. 
    \EndFor
    \State Reduce $g_b^{\rm risk}$ across row groups. 
    \State Aggregate modality norms and add the regularizer gradient in \cref{eq:reggradient}.
    \State Set $L_k\gets\widehat L_{k-1}$.
    \Repeat
        \State $\theta_{k+1}\gets u_k-L_k^{-1}\nabla J(u_k)$.
        \State Evaluate the trial objective by a score-only reduction. 
        \State $\mathrm{accepted}\gets[J(\theta_{k+1})\leq Q_{L_k}(\theta_{k+1},u_k)]$. \Comment{majorization test}
        \If{not $\mathrm{accepted}$}
            \State $L_k\gets\gamma L_k$.
        \EndIf
    \Until{$\mathrm{accepted}$}
    \State $\widehat L_k\gets L_k$; synchronize the accepted iterate. 
    \State $t_{k+1}\gets\{1+\sqrt{1+4t_k^2}\}/2$.
    \State $u_{k+1}\gets\theta_{k+1}+\frac{t_k-1}{t_{k+1}}(\theta_{k+1}-\theta_k)$. \Comment{momentum}
    \State Stop if $\norm{\theta_{k+1}-\theta_k}/(1+\norm{\theta_k})$ is below tolerance.
\EndFor
\State \Return $\widehat\theta$ and the modality weights.

\end{algorithmic}
\end{algorithm}

Algorithm~\ref{alg:2dfed} makes the resulting communication points explicit. Raw feature vectors are not transmitted: column workers exchange partial scores, sites exchange loss summaries, and row groups reduce feature-block gradients and the scalar intercept gradient. The score and risk reductions create synchronization barriers because $d_i$ depends on the complete additive score and $q_s$ depends on all site risks. The method is therefore synchronous, although local computation remains parallel along both partition axes.

Each failed line-search test increases $L_k$ and recomputes the trial point before testing again. Supplementary Material~S4.8 gives a gradient-norm objective-gap certificate when $\lambda_0,\eta_b>0$; the relative-iterate stopping rule used in the experiments does not supply an objective-gap certificate.

\subsection{Modality availability and propensity estimation}

Sample-level modality absence belongs to the statistical model, and it is distinct from system failure. Under the inverse-probability instantiation, an absent block has $r_{im}=a_{im}=0$ and contributes no raw value, while an observed block is reweighted by the clipped propensity estimate defined in \cref{sec:model}. Under the conditional-imputation instantiation, an absent optional block is replaced by a prediction from the always-observed block and enters with unit weight. The inverse-probability form is a modality-score analogue of correction for incomplete data \cite{little2019statistical,robins1994estimation}. The distinction matters in multimodal federated learning, where clients may legitimately hold different modality subsets \cite{huang2024multimodal}. Statistically absent sample--modality contributions are skipped in the parallel loops without altering the row/column reduction pattern.

For each optional modality, we fit a logistic observation model using the always-observed baseline block and site indicators. The required gradients are aggregated horizontally across row groups. If the baseline block and the availability indicator are held by different feature custodians, the logistic linear predictor can be computed from the same partial scores. Propensity models are fitted on training samples only and then applied unchanged to test samples. More flexible calibrated classifiers may replace logistic regression, but the clipping threshold should be specified in advance.

\subsection{Communication and system considerations}

Communication is a primary design constraint in federated learning \cite{kairouz2021advances,mcmahan2017communication}, and hybrid sample--feature partitioning creates traffic along both the row and column dimensions \cite{gao2023fedhd,zhang2024hybrid}. Let $n$ be the total number of aligned training rows and $d$ the total feature dimension. With 64-bit scalars and synchronous full gradients, the payload is approximately
\begin{equation}
    \operatorname{Comm}_{A,B}=16dA+16n(B-1)\quad\text{bytes per round}.
    \label{eq:communication}
\end{equation}
The first term accounts for gradient reduction and parameter broadcast across row blocks, at eight bytes in each direction; the second accounts for partial-score aggregation and derivative return across column blocks. Site-risk and modality-norm summaries add $O(S+M)$ scalars. Equation~\eqref{eq:communication} is a collapsed-coordinator approximation for one complete score-and-gradient evaluation. Supplementary Material~S3 separately counts the explicit messages of a global coordinator, row coordinators, and feature workers; that topology also includes communication for $A=B=1$, so its totals are not interchangeable with this approximation, and the payload figures reported there are computed from that topology rather than from \cref{eq:communication}. Each backtracking trial, including the accepted one, requires an additional score-only objective pass, and rejected trials require further passes. Total training traffic consequently depends on the observed number of line-search trials, reported in the Supplementary Material.

The trust model assumes non-colluding parties and does not provide a formal privacy guarantee. Row coordinators hold labels and site identifiers; a global coordinator receives site-risk summaries, adaptive site weights, feature-block gradients and modality norms; and a feature worker holds one column block with its availability indicators. Raw features and labels are not transmitted, but partial scores and gradients carry information about the records. Exact reductions preserve the equivalence in \Cref{thm:equivalence}; secure aggregation with fixed-point encoding or quantization would require a separate error analysis.

In a balanced layout, each row group performs $O(nd/A)$ work. If both rows and feature coordinates are balanced and all $AB$ workers operate concurrently, worker $(a,j)$ performs approximately $O\!\left(nd/(AB)\right)$ local operations per full score-and-gradient evaluation. Synchronization, network latency, and stragglers limit the corresponding reduction in wall-clock time. Nonlinear random features increase $d$ but preserve the additive communication interface. This calculation quantifies application-level payload and excludes any secure-aggregation overhead.

\section{Theoretical properties}
\label{sec:theory}

We state the results for fixed, finite-dimensional feature maps and the parameter domains specified in \cref{sec:model}. The statistical bounds concern the binary loss. Deterministic multiclass properties and complete proofs appear in Supplementary Material~S4. \Cref{thm:accelerated} adapts accelerated analysis to the proposed objective, \Cref{thm:equivalence} establishes exact equivalence for the implementation, and \Cref{thm:generalization} with \Cref{cor:propensity} bounds the corrected-score worst-site risk.

\begin{assumption}[Bounded corrected features]
\label{ass:features}
For every $m$, $\norm{\phi_m(x)}_2\leq\kappa_m$. The true propensities $p_{sim}=p_m(z_{si},s)$ satisfy $p_m(z,s)\geq p_{\min}>0$, and any estimated propensities used by the estimator are clipped at $p_{\min}$. For each fixed correction map considered below, the augmented availability-corrected feature, including the intercept coordinate, satisfies $\norm{\widetilde\phi_{si}}_2\leq B_x$.
\end{assumption}

\begin{assumption}[Sampling and parameter class]
\label{ass:parameters}
For the generalization result, $\sum_m\norm{w_m}_2\leq B_w$ and $|b|\leq B_b$. Observations are independent and identically distributed within each site, and the site samples are mutually independent. The feature maps and true propensity functions are fixed independently of the sampled observations. For each modality, $r_{sim}\perp x_{sim}\mid(z_{si},s)$, equivalently $\Prob(r_{sim}=1\mid x_{sim},z_{si},s)=p_m(z_{si},s)$.
\end{assumption}

\subsection{Convexity and smoothness}

\begin{proposition}
\label{prop:smooth}
Under \Cref{ass:features}, the objective $J$ in \cref{eq:objective} is convex and continuously differentiable. On
$\mathcal B_{R_w}=\{(w,b):\norm{w}_2\leq R_w\}$, its gradient is Lipschitz with the valid bound
\begin{equation}
\begin{split}
    \overline L_J(R_w)={}&\frac{\tau B_x^2}{4}+\frac{B_x^2}{\rho}
    +\lambda_0\max(1,\eta_b)\\
    &+\lambda_1\left(2M+\frac{\sqrt{M}R_w}{\epsilon}\right).
\end{split}
\label{eq:lipschitz_bound}
\end{equation}
If $\lambda_0>0$ and $\eta_b>0$, then $J$ is strongly convex with modulus at least $\mu=\lambda_0\min(1,\eta_b)$.
\end{proposition}

In \cref{eq:lipschitz_bound}, the first term is contributed by the softplus margin loss and the second by the covariance component of the Hessian of the log-sum-exp site aggregate. The last term is local because the Hessian of $G_\epsilon^2/2$ grows with $\norm{w}$. Decreasing $\rho$ therefore produces a closer approximation to the worst-site risk but also increases curvature, creating an optimization--robustness trade-off.

\begin{theorem}[Accelerated convergence]
\label{thm:accelerated}
Let $\theta^\star$ minimize $J$. Fix $T\geq1$ and suppose $\mathcal B_{R_w}$ contains $\theta^\star$ and every extrapolated and trial point evaluated through round $T$. Let \Cref{alg:2dfed} use backtracking factor $\gamma>1$, initial estimate $\widehat L_{-1}>0$, and the nondecreasing accepted estimates specified in the algorithm. Define
\begin{equation*}
    L_{R_w}=\max\{\widehat L_{-1},\gamma\overline L_J(R_w)\}.
\end{equation*}
Then every accepted estimate is at most $L_{R_w}$, and
\begin{equation}
    J(\theta_T)-J(\theta^\star)
    \leq \frac{2L_{R_w}\norm{\theta_0-\theta^\star}_2^2}{(T+1)^2}.
    \label{eq:accelerated_rate}
\end{equation}
When $\mu>0$, restarting the acceleration every
$K=\lceil\sqrt{8L_{R_w}/\mu}\rceil$ gradient rounds reduces the objective gap by at least one half per epoch, provided the same regional bound holds for those epochs.
\end{theorem}

The $O(T^{-2})$ bound in \cref{eq:accelerated_rate} requires the stated local-region condition but not strong convexity. Strong convexity is needed only for the restart contraction. The theorem therefore assumes that extrapolated and rejected trial points remain in the stated region.

The constants in \cref{eq:lipschitz_bound} are explicit but loose at this scale. With $\epsilon=10^{-4}$, the group term is $\lambda_1(2M+\sqrt{M}R_w/\epsilon)$. Taking $R_w=26$, the bound implied by \Cref{cor:parameter_bounds} for the public settings, that term alone is about $1.8\times10^{3}$, while $\mu=6\times10^{-4}$. The restart epoch of \Cref{thm:accelerated}, $K=\lceil\sqrt{8L_{R_w}/\mu}\rceil$, is therefore at least about $4.9\times10^{3}$ rounds, above the 1400-round cap used in the experiments. The iteration counts in Section~\ref{sec:public} therefore reflect local curvature rather than the worst-case rate.

\subsection{Equivalence of two-dimensional and centralized gradients}

\begin{theorem}[Exact block-gradient equivalence]
\label{thm:equivalence}
Suppose $\{\Ical_a\}_{a=1}^A$ and $\{\mathcal J_j\}_{j=1}^B$ are disjoint partitions of the training samples and feature coordinates, respectively, and all reductions in \Cref{alg:2dfed} are exact. Then, for every $\theta$,
\begin{equation*}
\begin{aligned}
    \nabla_{w_j}J(\theta)
      &=\sum_{a=1}^{A}g_{aj}
        +\nabla_{w_j}\operatorname{Reg}(\theta),
        &&j=1,\ldots,B,\\
    \nabla_bJ(\theta)
      &=g_b^{\rm risk}
        +\frac{\partial\operatorname{Reg}(\theta)}{\partial b}.
\end{aligned}
\end{equation*}
Thus, with identical initialization and line-search decisions,
\begin{equation*}
    \theta_k^{\rm dist}=\theta_k^{\rm cent},
    \qquad J(\theta_k^{\rm dist})=J(\theta_k^{\rm cent}),
    \quad k=0,\ldots,T.
\end{equation*}
\end{theorem}

The identity holds at the level of individual iterates: the two-dimensional layout changes where sums are evaluated, but not the estimator or update being computed. In floating-point arithmetic, different reduction orders produce rounding-level deviations; \Cref{sec:partition_results} reports their observed magnitude. The theorem is related to the \emph{partition-insensitive} property studied by Wu et al.~\cite{wu2025dantzig}. For a feature-splitting proximal-point algorithm for Dantzig selectors, they showed that the terminal solution is unchanged by the partition of the feature set. Both results express partition invariance, but at different levels. \Cref{thm:equivalence} is an iterate-level statement: for any admissible row/column partition, exact synchronous aggregation makes every distributed gradient, accepted line-search value, and iterate coincide with its centralized counterpart. Invariance of the final estimator follows immediately. This operational result relies on the additive score decomposition and the synchronization and exact-reduction conditions in \Cref{thm:equivalence}. Stale, lossy, or asynchronous updates require a separate equivalence and convergence analysis.

\begin{corollary}[Parameter bounds for the regularized estimator]
\label{cor:parameter_bounds}
Assume $\lambda_0>0$ and $\eta_b>0$, and let $(\widehat w,\widehat b)$ minimize \cref{eq:objective}. Define
\[
J_0=\ell_\tau(0)+\frac{\lambda_1}{2}M^2\epsilon^2.
\]
Since $J(\widehat w,\widehat b)\leq J(0,0)=J_0$, the estimator satisfies
\[
\sum_{m=1}^M\|\widehat w_m\|_2\leq \sqrt{\frac{2M J_0}{\lambda_0}},
\qquad
|\widehat b|\leq \sqrt{\frac{2J_0}{\lambda_0\eta_b}}.
\]
These choices of $B_w$ and $B_b$ establish only the parameter restrictions in \Cref{ass:parameters}; its sampling, fixed-map, and missingness assumptions remain separate requirements.
\end{corollary}

The corollary concerns an exact minimizer. A numerical output with $J(\theta)\leq J_0+e$, $e\geq0$, satisfies the same bounds with $J_0+e$ in place of $J_0$. Supplementary Material~S4.8 gives a gradient-norm gap certificate for runs that record the final gradient evaluation.

\subsection{Worst-site generalization under modality absence}

For any fixed propensity map $\bar p_m(z,s)\geq p_{\min}$, write
\begin{equation*}
    f^{\bar p}_{w,b}(x,r,z,s)
    =b+\sum_{m=1}^{M}\frac{r_m}{\bar p_m(z,s)}
      \ip{w_m}{\phi_m(x_m)},
\end{equation*}
and let $R_s^{\bar p}(f)$ and $\widehat R_s^{\bar p}(f)$ denote its population and empirical soft-margin risks at site $s$. The oracle bound is stated for a fixed true propensity map $p$; treating a nuisance estimate fitted on the same data requires separate estimation control. The proof uses a standard Rademacher-complexity deviation inequality \cite{bartlett2002rademacher}. Let $n_{\min}=\min_sn_s$, $\pi_{\min}=\min_s\pi_s$, and $\kappa=(\sum_m\kappa_m^2)^{1/2}$. The parameter bounds give the loss envelope
\begin{equation*}
    L_{\max}=\ell_\tau\!\left(-B_b-B_w\kappa/p_{\min}\right).
\end{equation*}

\begin{theorem}[Uniform oracle-propensity worst-site bound]
\label{thm:generalization}
Under \Cref{ass:features,ass:parameters}, with probability at least $1-\delta$, simultaneously for every admissible $f$,
\begin{equation}
\begin{split}
    \max_{s\in\Scal}R_s^p(f)
    \leq{}&\softmaxrisk\{\widehat R_1^p(f),\ldots,\widehat R_S^p(f)\}
    +\rho\log\frac{1}{\pi_{\min}}\\
    &+\frac{2(B_w\kappa/p_{\min}+B_b)}{\sqrt{n_{\min}}}
    +3L_{\max}\sqrt{\frac{\log(2S/\delta)}{2n_{\min}}}.
\end{split}
\label{eq:generalization}
\end{equation}
\end{theorem}

The bound concerns binary soft-margin risk for the oracle-corrected, partially observed score. Extending it to the learned multiclass pipeline with training-fitted PCA and propensity models requires additional analysis. Rarely observed modalities inflate the complexity term through $1/p_{\min}$; the smallest site determines the worst-site sampling term; and the reference weights contribute $\rho\log(1/\pi_{\min})$.

\begin{corollary}[Estimated-propensity perturbation]
\label{cor:propensity}
Let $\widehat p$ denote the post-clipping propensity estimator used in the empirical objective. Suppose the event
$\mathcal E_p=\{\max_{s,i,m}|\widehat p_{sim}-p_{sim}|\leq e_p\}$ has probability at least $1-\delta_p$. On this event, replacing $p$ by $\widehat p$ changes every admissible training loss by at most
\begin{equation*}
    \Delta_p=\frac{B_w\kappa e_p}{p_{\min}^2}.
\end{equation*}
Consequently, for $\delta+\delta_p<1$, with probability at least $1-\delta-\delta_p$, the bound in \cref{eq:generalization} remains valid after replacing
$\softmaxrisk(\widehat R^p)$ by
$\softmaxrisk(\widehat R^{\widehat p})+\Delta_p$.
\end{corollary}

Because the comparison is pathwise on $\mathcal E_p$, the corollary allows the propensity model to be fitted on the training sample. The result bounds perturbation of the empirical objective relative to the oracle-propensity population target. A usable high-probability value of $e_p$ and out-of-sample control of fitted propensity functions are separate estimation problems. Across the primary masks the fitted propensities deviate from the true ones by $e_p=0.08$--$0.16$ on MHEALTH and $0.02$--$0.15$ on HAR; with $B_w$ from \Cref{cor:parameter_bounds}, the observed feature norms and $p_{\min}=0.08$, the perturbation constant is $\Delta_p\approx1.1\times10^{4}$--$2.3\times10^{4}$ on MHEALTH and $0.9\times10^{4}$--$6.0\times10^{4}$ on HAR (Supplementary Material~S2). At this scale $\Delta_p$ exceeds the loss range, so the corollary is qualitative. Under missing-not-at-random mechanisms, availability may still depend on unobserved modality values after conditioning on $z$, requiring additional identification assumptions. The controlled sensitivity experiments therefore examine the correction under specified departures from the default MAR model.

\section{Simulation study}
\label{sec:simulations}

\subsection{Design and comparators}

The simulation isolates three features of the proposed setting: blockwise modality absence, unequal sites with distributional shifts, and row/column distributed optimization. We use $S=3$ sites and $M=4$ modalities of dimensions 8, 12, 16, and 12. The site-specific training sizes are $(360,140,60)$, and each test site contains 300 samples. A five-dimensional latent state $z_{si}\sim N(h\mu_s,I_5)$ generates the modality measurements and labels. The heterogeneity parameter $h$ controls shifts in both the site distributions and the discriminant directions, with the largest decision-boundary shift assigned to the smallest site in the primary design. This assignment combines the smallest sample with the largest shift, the configuration in which worst-site aggregation has most to gain; Section~\ref{sec:design} reports reversed and size-balanced designs as a check on the mechanism.

Modality 1 is always observed. Each optional modality follows the bounded MAR mechanism
\begin{equation}
 p_{sim}=\operatorname{clip}\!\left[\sigma\{\alpha_m+\gamma_{sm}+0.35x_{si1}^{(1)}-0.25x_{si2}^{(1)}+0.15x_{si3}^{(1)}\},0.08,0.98\right],
 \label{eq:simmissing}
\end{equation}
where $\sigma$ is the logistic function and the modality-specific site offsets $\gamma_{sm}$ are specified in Supplementary Material~S1. Propensities are re-estimated on the training data by ridge-logistic regression and clipped to $[0.10,0.98]$. We study $h\in\{0,0.4,0.8,1.2\}$ at 35\% target optional-modality absence and missingness levels $\{0.10,0.25,0.35,0.50,0.65\}$ at $h=0.8$, with 20 matched replicates per scenario. Hyperparameters are fixed across scenarios: $\tau=4$, $\rho=0.06$, $\lambda_0=0.015$, $\lambda_1=0.004$, $\epsilon=10^{-4}$, and $\eta_b=0.05$.

We compare Complete-case SMKL, Zero-fill SMKL, IPW mean SMKL, 2D-FedSMKL, and an infeasible Oracle full SMKL reference. Unless stated otherwise, every smooth baseline uses the same group-kernel penalty as the proposed method. Worst-site AUC is the primary metric; mean-site AUC and mean and worst-site balanced accuracy are secondary metrics. A worst-site value is the minimum over sites of the metric in question, and the minimizing site can differ across metrics. Because each modality is standardized on its observed training rows, zero filling places an absent block at its observed training mean, so the zero-fill baseline is mean imputation. Supplementary Material~S1 specifies the data-generating process and reports paired intervals and stress diagnostics; the project repository retains the replicate-level records. Complete-case training uses zero-filled inputs for inference on the same masked test population.

The archived simulations used Python 3.13.5 and NumPy 2.3.5 on a Linux CPU instance. The public-data and explicit-message experiments use Python 3.12.14, NumPy 2.3.5, SciPy 1.18.1 and scikit-learn 1.9.1 on Windows x86\_64 with one BLAS thread. Wall-clock measurements come from single-process emulations and exclude physical-network latency and concurrent worker execution. Exact versions, input hashes and commands are recorded with each run. Reproduction materials and deterministic seeds are available at \url{https://github.com/xfwu1016/2D-FedSMKL}.

\subsection{Primary predictive results}

\begin{table*}[t]
\centering
\caption{Predictive performance in the primary simulation setting (35\% target optional-modality absence and heterogeneity $h=0.8$). Entries are mean (standard deviation) over 20 replicates. The oracle uses fully observed modalities and is not feasible in practice.}
\label{tab:primary}
\scriptsize
\begin{tabular}{lcccc}
\toprule
Method & Mean AUC & Worst-site AUC & Mean BA & Worst-site BA \\
\midrule
Complete-case SMKL & 0.675 (0.024) & 0.583 (0.030) & 0.621 (0.025) & 0.546 (0.027) \\
Zero-fill SMKL & 0.740 (0.018) & 0.613 (0.034) & 0.673 (0.016) & 0.574 (0.032) \\
IPW mean SMKL & 0.736 (0.018) & 0.612 (0.035) & 0.669 (0.014) & 0.577 (0.029) \\
\textbf{2D-FedSMKL} & 0.723 (0.022) & 0.622 (0.037) & 0.659 (0.018) & 0.584 (0.028) \\
Oracle full SMKL & 0.751 (0.016) & 0.623 (0.037) & 0.681 (0.015) & 0.581 (0.036) \\
\bottomrule
\end{tabular}
\end{table*}

In the primary setting ($h=0.8$, 35\% target absence), 2D-FedSMKL attains a worst-site AUC of 0.622, compared with 0.612 for IPW mean SMKL and 0.583 for complete-case training (\Cref{tab:primary}). Its mean AUC is lower than that of IPW mean SMKL (0.723 versus 0.736). The robust criterion gives more weight to the most difficult site rather than maximizing prevalence-weighted performance; the lower mean AUC is the cost of that choice. In one prespecified reference replicate, the fitted modality weights are 0.398, 0.157, 0.230, and 0.215; the group-kernel penalty retains all four modalities rather than selecting a single block.

The heterogeneity stress test shows when this trade-off becomes relevant. At $h=1.2$, the worst-site AUC of 2D-FedSMKL exceeds that of IPW mean SMKL by 0.049 (95\% paired interval 0.039--0.059), whereas the two objectives give almost identical results at $h=0$. The effect of robust aggregation thus grows when one site becomes systematically more difficult, but is negligible when site-specific decision rules are similar. This pattern agrees with Eq.~\eqref{eq:siteweight}: a site with a larger empirical loss has greater influence on the gradient than it would under prevalence weighting.

Increasing missingness produces a different pattern. The number of complete records falls rapidly, making complete-case fitting unstable: at 65\% target absence, only 6 of 20 replicates meet the prespecified fitting rule. Because available modality scores from incomplete records still contribute, 2D-FedSMKL remains estimable. At this extreme level, however, its worst-site AUC exceeds that of IPW mean SMKL by only 0.009 and remains close to that of zero filling. Inverse weighting can have high variance when estimated propensities are small or variable, so unbiased score contributions in expectation need not yield better predictions in a finite sample. The missingness experiment supports retaining incomplete records, while the finite-sample benefit of propensity weighting varies with the missingness level and propensity dispersion.

\begin{figure}[t]
\centering
\includegraphics[width=0.97\textwidth]{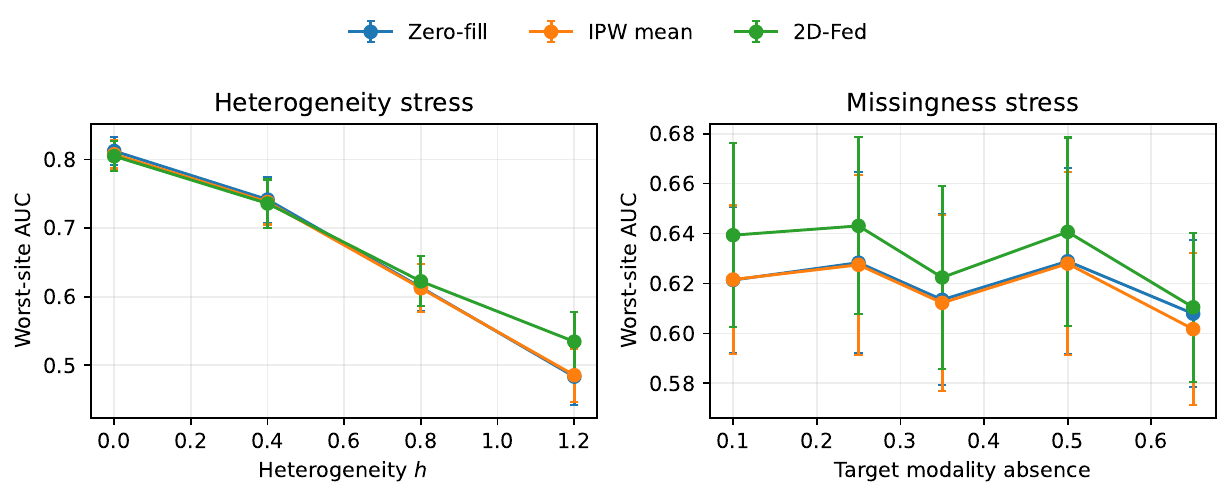}
\caption{Compact simulation trends for the three incomplete-sample fusion methods. The legend label 2D-Fed denotes the proposed 2D-FedSMKL method. Left: the robust objective separates from mean-risk alternatives as site heterogeneity increases. Right: the methods remain close at moderate modality absence, with only modest separation at severe absence. Error bars show standard deviations across matched replicates.}
\label{fig:simtrends}
\end{figure}

\subsection{Optimization and partition audit}
\label{sec:partition_results}

Across the tested row/column layouts, the explicit two-dimensional and centralized gradients differ by between $4.16\times10^{-17}$ and $1.04\times10^{-16}$, consistent with \Cref{thm:equivalence}. In the reference fit, the smallest site represents $60/560=0.107$ of the training sample but receives a final adaptive weight of 0.303 because it has the largest empirical risk. Supplementary Material~S1 provides the objective trace, partition table, communication-by-layout audit, and summaries of the 786 successful fits; the replicate-level records are retained in the project repository. The remaining 14 results are failed complete-case fits at the highest missingness level.

The multiclass message interface is also exercised on both public datasets, using separate row--column feature arrays and explicit score, derivative, norm and gradient messages. Across $(A,B)\in\{(1,1),(2,2),(4,4)\}$, objective and gradient checks agree with the centralized evaluator to below $10^{-12}$, and each solver gives the same test predictions as its centralized counterpart. With a common fixed objective and a 350-round budget, FISTA reaches a $10^{-3}$ gap to an independently optimized numerical reference after 188 rounds on MHEALTH and 204 on HAR. Plain gradient descent does not reach that target within the budget. We use the best observed objective as a numerical reference; it is not treated as the exact optimum $J^\star$. Supplementary Material~S3 gives the message ledger, terminal gradient norms, reference certificate and complete accuracy--payload traces.
\subsection{Design sensitivity of the worst-site gain}
\label{sec:design}

The primary design assigns the largest shift to the smallest site. Repeating the simulation with the largest shift assigned to the largest site leaves the worst-site contrast at $h=1.2$ at $+0.018$ $[+0.009,+0.028]$ and removes it at $h=0.8$ ($0.000$ $[-0.004,+0.004]$). With site sizes $(187,187,186)$, the contrast is $+0.014$ $[+0.005,+0.022]$ at $h=0.8$ and $+0.025$ $[+0.014,+0.035]$ at $h=1.2$. The mechanism therefore does not depend on the smallest site being the hardest, or on unequal site sizes; its magnitude does. Supplementary Material~S1 reports the full design comparison.

\section{Public-data multiclass validation}
\label{sec:public}

\subsection{Datasets, representations and source-only validation}
MHEALTH provides wearable signals from 10 participants performing 12 activities \cite{banos2014mhealthdata,banos2014mhealthdroid}. We extract nonoverlapping 2-s windows and use four blocks (chest acceleration, ECG, ankle and wrist sensing), retaining 55 coordinates after training-fitted reduction. Each participant is a statistical site. Within participant/activity, the earliest 70\% of windows train the model and the remainder test it (2335/1020 windows); this evaluates later windows from seen participants. UCI HAR provides six smartphone activities from 30 volunteers \cite{reyesortiz2013ucidata,anguita2013public}. We retain its official subject-disjoint 7352/2947-window split, assign subjects to five fixed silos, and reduce accelerometer/non-gyroscope and gyroscope blocks to 25 PCA coordinates each. Test subjects are held out, while the five silos are constructed subject groups. MHEALTH therefore has 10 participant sites and HAR has five constructed silos; the two definitions follow the available grouping variables and are not equivalent.

Both datasets are originally complete. We impose site- and baseline-dependent MAR absence, keep the first block observed, and fit observation propensities on training masks. Primary evaluation uses 35\% target optional-block absence and ten matched masks. Scaling and PCA are fitted exclusively within each inner-training fold for validation, then refitted on the full source-training partition for outer testing. HAR uses raw-block scaling followed by PCA, without subsequent PCA-coordinate scaling. Standardization and PCA are applied block by block, so a custodian that holds one block can compute them locally; they are fitted here on complete source modalities before controlled removal, so the benchmark evaluates decision fusion conditional on those maps and does not model representation learning under missingness. Because the observed blocks are standardized, zero filling is mean imputation on these benchmarks. The theoretical risk bound covers binary learning with fixed maps; the benchmark tests the multiclass extension empirically.

The source-only grids select $\rho=0.12$ for MHEALTH and $0.20$ for HAR, SVM $C=10$ and 1, and FedAvg/FedProx learning rate 0.05 with 350 and 200 rounds, respectively. Five local steps and FedProx coefficient 0.1 are fixed. All final primary smooth models and the robustness grid use a uniform 1400-round cap after training-objective/gradient diagnostics of the 350-round pilot. Other smooth parameters are fixed. Supplementary Material~S2 specifies their values, the folds, candidate grids, selection rule, seeds and numerical diagnostics. Model selection uses inner validation only; outer test labels are reserved for final evaluation.

\subsection{Comparators and primary results}
References are pooled linear SVM \cite{cortes1995support}, complete-case smooth Crammer--Singer (CS) classification \cite{little2019statistical,crammer2001multiclass,lee2001ssvm}, FedAvg-CS \cite{mcmahan2017communication}, FedProx-CS \cite{li2020federated}, Zero-fill SMKL \cite{rakotomamonjy2008simplemkl}, IPW mean SMKL \cite{rakotomamonjy2008simplemkl,robins1994estimation}, and an infeasible full-modality oracle. Complete-case CS trains on fully observed rows with no MKL penalty and uses zero-filled inference on the common masked test population. Its comparison jointly reflects training-row selection, regularization, and test-time zero filling. The centralized and horizontal methods are predictive references; the separate feature-custodian constraint is assessed by the row/column message-interface experiment. The same-interface computational control is reported in Section~6 and Supplementary Material~S3. External modality-completion, hallucination and distillation baselines are not included; the conditional-imputation instantiation is part of the method and is compared in Section~\ref{sec:correction}. The remaining comparison isolates training-row selection, mean imputation, availability correction, aggregation and horizontal federated training.

\begin{table}[!t]
\centering
\caption{Public-data results with fold-fitted preprocessing for source validation and final source-only representations, under 35\% controlled MAR block absence. Entries are mean (sample SD) over ten matched masks. Zero-fill SMKL, and the zero-filled inference of complete-case CS, place an absent block at its observed training mean, which is mean imputation on standardized blocks. The oracle uses unmasked data and is fitted once, so its SD is zero across masks.}\label{tab:public_main}
\scriptsize
\setlength{\tabcolsep}{4pt}
\begin{tabular}{lcccc}
\toprule
Method & Mean macro-F1 & Worst-site macro-F1 & Mean BA & Worst-site BA \\
\midrule
\multicolumn{5}{l}{\textbf{MHEALTH}} \\
Pooled linear SVM & 0.949 (0.007) & 0.849 (0.025) & 0.949 (0.007) & 0.859 (0.027) \\
Complete-case CS & 0.763 (0.021) & 0.587 (0.060) & 0.767 (0.021) & 0.592 (0.062) \\
FedAvg-CS & 0.925 (0.012) & 0.856 (0.029) & 0.925 (0.012) & 0.858 (0.028) \\
FedProx-CS & 0.924 (0.012) & 0.856 (0.029) & 0.925 (0.012) & 0.858 (0.028) \\
Zero-fill SMKL & 0.927 (0.011) & 0.856 (0.032) & 0.927 (0.011) & 0.855 (0.033) \\
IPW mean SMKL & 0.924 (0.009) & 0.850 (0.032) & 0.925 (0.009) & 0.850 (0.030) \\
\textbf{2D-FedSMKL (IPW)} & 0.921 (0.013) & 0.857 (0.027) & 0.922 (0.013) & 0.860 (0.028) \\
Oracle full SMKL & 0.988 (0.000) & 0.953 (0.000) & 0.988 (0.000) & 0.954 (0.000) \\
\midrule
\multicolumn{5}{l}{\textbf{UCI-HAR}} \\
Pooled linear SVM & 0.885 (0.005) & 0.772 (0.012) & 0.885 (0.005) & 0.775 (0.011) \\
Complete-case CS & 0.858 (0.006) & 0.775 (0.012) & 0.862 (0.005) & 0.787 (0.011) \\
FedAvg-CS & 0.881 (0.003) & 0.788 (0.009) & 0.881 (0.003) & 0.790 (0.009) \\
FedProx-CS & 0.881 (0.003) & 0.788 (0.009) & 0.881 (0.003) & 0.790 (0.009) \\
Zero-fill SMKL & 0.884 (0.003) & 0.777 (0.006) & 0.884 (0.003) & 0.781 (0.006) \\
IPW mean SMKL & 0.885 (0.004) & 0.787 (0.010) & 0.885 (0.004) & 0.790 (0.010) \\
\textbf{2D-FedSMKL (IPW)} & 0.884 (0.004) & 0.787 (0.010) & 0.884 (0.004) & 0.790 (0.010) \\
Oracle full SMKL & 0.908 (0.000) & 0.826 (0.000) & 0.907 (0.000) & 0.829 (0.000) \\
\bottomrule
\end{tabular}
\end{table}

Table~\ref{tab:public_main} reports the inverse-probability instantiation; the validation-selected configuration is reported in Section~\ref{sec:correction}.

On MHEALTH, mean worst-site macro-F1 is 0.857 for the IPW instantiation of 2D-FedSMKL, 0.856 for Zero-fill (mean-imputation) SMKL, 0.850 for IPW mean SMKL, and 0.856 for FedAvg-CS. The paired differences are small and their intervals include zero: $[-0.010,+0.024]$ against IPW mean SMKL and $[-0.017,+0.018]$ against FedAvg-CS. The intervals therefore bound a possible difference rather than establish one. The oracle in \Cref{tab:public_main} is fitted once on unmasked data, so its standard deviation across masks is zero.

On HAR, the corresponding worst-site values are 0.787, 0.777, 0.787, and 0.788. The IPW instantiation improves on mean imputation by 0.010 (95\% interval $[+0.006,+0.014]$), but its intervals against IPW mean SMKL, $[-0.002,+0.002]$, and against FedAvg-CS, $[-0.006,+0.003]$, include zero. The gain over mean imputation therefore does not extend to the mean-risk MKL or horizontal references on this benchmark. The paired differences and their 95\% mask-conditional intervals are reported in Supplementary Material~S2. These intervals describe variation across controlled masks on fixed samples and representations, without multiplicity adjustment; they do not extend to new sites or the population.

\subsection{Component stress tests and numerical scope}
A separate prespecified $2\times2$ study crosses mean/robust aggregation with zero-fill/IPW at 10\%, 35\% and 65\% absence, using 20 matched masks per level, a common $\rho=0.05$, and a 350-round budget. Figure~\ref{fig:publicstress} displays the paired component contrasts and their unadjusted 95\% intervals at 35\% and 65\% absence. For robust aggregation under IPW, the paired gains are +0.0157 and +0.0196 on MHEALTH at 35\% and 65\% absence, respectively, but only +0.0006 and +0.0038 on HAR. The point estimates are larger on MHEALTH and grow with absence there, but the intervals are unadjusted and the study contains 24 contrasts, so we read these contrasts as descriptive. For IPW relative to zero filling under robust aggregation, the corresponding differences are $-0.0022$ and +0.0004 on MHEALTH, compared with +0.0056 and +0.0080 on HAR. The estimated interaction is small relative to its uncertainty, so robust aggregation and IPW correction show no clear joint effect. Full paired intervals, including the mean-risk contrasts, are provided in Supplementary Material~S2.

The primary protocol and the component study use different temperatures. Validation selects $\rho=0.12$ for MHEALTH and $0.20$ for HAR, values at which the log-sum-exp aggregate is close to mean aggregation; the primary robust-versus-mean contrasts are correspondingly small. The component study uses $\rho=0.05$, where the robust term carries more influence. The component study therefore shows that the mechanism operates at smaller $\rho$; it does not show that the selected operating point improves worst-site performance. Re-running the validation grid on ten masks at the full 1400-round budget selects $\rho=0.08$ for MHEALTH ($0.8169$, against $0.8168$ at $0.12$) and $\rho=0.20$ for HAR (Supplementary Material~S2); the selection is therefore stable, and the smaller-$\rho$ gain seen on the test masks is not selected by validation. The selection is stable, and the smaller-$\rho$ gain seen on the test masks is not selected by validation.

The effect of MKL is dataset-dependent: on MHEALTH at 65\% absence, removing the group-kernel penalty raises worst-site macro-F1 from 0.787 to 0.806 (full minus no-MKL: $-0.0191$, interval $[-0.0289,-0.0093]$), whereas the corresponding HAR difference is small (+0.0019). The paired intervals are reported in Supplementary Material~S2. Coefficient-norm shares summarize fitted modality weights and are interpreted descriptively. Propensity clipping and deliberate covariate omission are audited on unchanged masks in S2. All 620 secondary fits reach their 350-round cap. Their terminal-gradient diagnostics and the primary three-mask 350/1400-round budget comparison are reported in Supplementary Material~S2; the component results are therefore interpreted as fixed-budget comparisons.

\subsection{Validation-selected correction}
\label{sec:correction}

The correction is selected on validation data. For each dataset, both instantiations are fitted on the inner-training partition at the temperature validated for the inverse-probability instantiation ($\rho=0.12$ for MHEALTH, $0.20$ for UCI-HAR; 1400 rounds) and compared on ten inner-validation masks (seeds 20261131--20261140; Supplementary Material~S2). Conditional imputation is selected on both: $0.8282$ against $0.8168$ on MHEALTH and $0.5769$ against $0.5549$ on UCI-HAR. The selection uses validation data only; the test masks play no part.

Table~\ref{tab:impute} reports the selected model on the ten test masks. On MHEALTH it attains $0.882$, against $0.856$ for mean imputation in \Cref{tab:public_main}, $0.857$ for the IPW instantiation and $0.850$ for IPW mean. The paired differences are $+0.026$ $[-0.000,+0.052]$ against mean imputation, $+0.032$ $[+0.008,+0.056]$ against IPW mean, $+0.025$ $[+0.008,+0.043]$ against the IPW instantiation, and $+0.026$ $[+0.000,+0.052]$ against the horizontal FedAvg reference. On UCI-HAR the selected model attains $0.787$; the intervals against IPW mean, the IPW instantiation and FedAvg-CS are $[-0.009,+0.008]$, $[-0.009,+0.008]$ and $[-0.008,+0.005]$, while the differences against mean imputation, $+0.010$ $[+0.001,+0.018]$, and against the pooled linear SVM, $+0.015$ $[+0.007,+0.022]$, exclude zero.

The correction matters: validation prefers imputation on both benchmarks, and on MHEALTH the selected model is significantly better than the IPW instantiation and IPW mean; the difference from the FedAvg reference is borderline, with interval $[+0.000,+0.052]$. The margin over mean imputation is positive but borderline on MHEALTH ($[-0.000,+0.052]$) and significant on UCI-HAR, so the selected model is competitive but not uniformly superior. The risk analysis of Section~\ref{sec:theory} covers the IPW instantiation, whose score remains unbiased under MAR. The selected imputation instantiation is evaluated empirically; its risk is not covered by that bound, and imputation requires the always-observed block at the missing block's custodian, whose communication and privacy costs are not analysed here.

Two further references were run at the 350-round budget, each on the masks of the comparison it supports. A modality-dropout training baseline (random block removal during training, three augmented copies, 20 masks) attains $0.861$ on MHEALTH and $0.772$ on UCI-HAR, below the selected model on the same masks ($0.884$ and $0.784$; Supplementary Material~S2). A late-fusion reference (one smooth classifier per block, scores averaged over observed blocks) attains $0.827$ on MHEALTH and $0.794$ on UCI-HAR on the ten primary masks; the selected model is $+0.056$ $[+0.039,+0.072]$ above it on MHEALTH and $-0.006$ $[-0.012,+0.000]$ on UCI-HAR, so the multiple-kernel layer contributes where several heterogeneous blocks are present and is indistinguishable when only two blocks are available.

% Table 3 and Figure 5 are intentionally composed on one float page.
\begin{table}[p]
\centering
\caption{Availability correction under the primary protocol (35\% target absence, ten masks, 1400 rounds). Worst-site macro-F1 and paired difference against IPW mean, with the 95\% interval over ten masks. The last row is the configuration selected on validation data and is the model summarised in the abstract.}
\label{tab:impute}
\footnotesize
\setlength{\tabcolsep}{3pt}
\begin{tabular}{lcc}
\toprule
Method & Worst-site macro-F1 & $\Delta$ vs IPW mean\\
\midrule
\multicolumn{3}{l}{\textbf{MHEALTH}}\\
IPW mean & 0.850 & -- \\
2D-FedSMKL (IPW, robust) & 0.857 & +0.007 [-0.010, +0.024] \\
Conditional imputation & 0.877 & +0.027 [+0.008, +0.047] \\
Conditional imputation + robust & 0.882 & +0.032 [+0.008, +0.056] \\
\midrule
\multicolumn{3}{l}{\textbf{UCI-HAR}}\\
IPW mean & 0.787 & -- \\
2D-FedSMKL (IPW, robust) & 0.787 & +0.000 [-0.002, +0.002] \\
Conditional imputation & 0.785 & -0.002 [-0.008, +0.004] \\
Conditional imputation + robust & 0.787 & -0.000 [-0.009, +0.008] \\
\bottomrule
\end{tabular}
\vspace{0pt}
\includegraphics[width=0.88\textwidth]{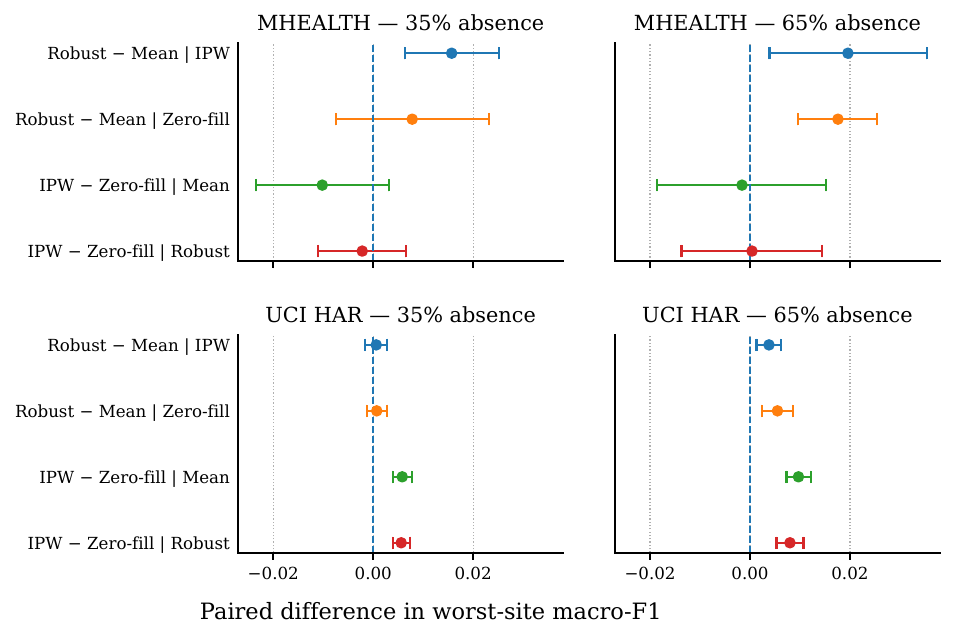}
\captionof{figure}{Paired component effects in the secondary factorial study under controlled MAR block absence. Panels correspond to MHEALTH and UCI HAR at 35\% and 65\% target absence. Points are paired mean differences in worst-site macro-F1 and horizontal bars are unadjusted 95\% paired $t$ intervals over 20 matched masks. Positive values favor the first-listed component: robust minus mean aggregation, conditional on IPW or zero filling; and IPW minus zero filling, conditional on mean or robust aggregation. The vertical dashed line denotes no difference. The figure focuses on the absence levels for which paired inferential contrasts are tabulated.}
\label{fig:publicstress}
\end{table}

\FloatBarrier % Keep Section 7 tables and figures out of the Discussion section.
\clearpage
\section{Discussion and conclusion}
\label{sec:discussion}

This paper develops 2D-FedSMKL, a hybrid federated decision layer for multimodal fusion with sample and feature partitions. For fixed representations the objective is convex and smooth; under the partition and exact-reduction conditions of Theorem~2 the distributed updates coincide with the centralized iterates; the curvature bounds and accelerated rate characterize optimization on the stated region; and the risk result is restricted to the binary, oracle-propensity setting. In the simulations, worst-site aggregation improves worst-site performance under heterogeneity, and the gain survives the reversed and size-balanced designs. On the public benchmarks the correction is selected on validation data, and the selected model is never significantly worse than the strongest reference at the worst site. On MHEALTH it is above mean imputation, the inverse-probability instantiation and the horizontal reference; on UCI-HAR the differences from the mean-risk and horizontal references include zero.

The component analyses qualify these gains and delimit the evidence. The correction matters more than the aggregation rule on MHEALTH: validation selects conditional imputation on both benchmarks, and the selected model is above the inverse-probability instantiation on MHEALTH. The group-kernel penalty helps on UCI-HAR and hurts MHEALTH at 65\% absence, and the robust term is visible only at the smaller temperature of the component study. Four limits bound the evidence: the masks are synthetic MAR generated by the same logistic form as the fitted model, so misspecified and missing-not-at-random mechanisms are untested; the representations are fitted on complete sources before masking; no distillation or hallucination baseline is included; and the risk analysis covers the inverse-probability instantiation rather than the selected one, whose imputation step also adds an unmodelled model-selection step. The communication audit is a single-process emulation, the trust model assumes non-colluding parties, and the algorithm is synchronous.

Future work will study naturally incomplete multimodal data, federated representation learning under missingness, and misspecified or missing-not-at-random propensities. A risk analysis for the imputation instantiation, a selection rule that accounts for validation uncertainty, and comparisons with distillation methods are the next methodological steps. An asynchronous implementation would require explicit delay assumptions and a separate convergence analysis.

\clearpage

\section*{CRediT authorship contribution statement}
\textbf{Xiaofei Wu:} Conceptualization, Methodology, Software, Validation, Formal analysis, Investigation, Data curation, Writing -- original draft, Visualization. \textbf{Rongmei Liang:} Data curation, Formal analysis, Investigation, Writing -- review \& editing. \textbf{Jian Qing Shi:} Conceptualization, Methodology, Supervision, Writing -- review \& editing, Funding acquisition, Project administration.

\section*{Declaration of competing interest}
The authors declare that they have no known competing financial interests or personal relationships that could have appeared to influence the work reported in this paper.

\section*{Declaration of generative AI and AI-assisted technologies in the manuscript preparation process}
OpenAI Codex assisted with text, proof drafts, experimental code, and figure-generation code. All authors have reviewed and approved the final manuscript.

\section*{Funding}
Jian Qing Shi's work was supported in part by the National Key R\&D Program of China (Grant No. 2023YFA1011400), the Guangdong Higher Education Upgrading Plan (Grant No. 0400008-26), and the BNBU Research Grant (Grant No. R700105-27).

\section*{Data and code availability}
The raw MHEALTH and Human Activity Recognition Using Smartphones data are publicly available from the UCI Machine Learning Repository under DOIs 10.24432/C5TW22 and 10.24432/C54S4K, respectively. No new clinical or participant data were collected for this study. Reproduction code, simulation outputs, deterministic seeds, derived feature caches, validation indices, sample-level primary predictions and raw benchmark result tables are available at the project repository \url{https://github.com/xfwu1016/2D-FedSMKL} (release v24.8). The repository identifies the exact revision supporting the present tables.

% Insert an Acknowledgements section here only if the authors have content to report.

\clearpage
\section*{Supplementary Material}
\noindent This supplement reports simulation stress tests, public-benchmark tuning and diagnostics, communication costs including backtracking objective passes, and proofs accompanying the hybrid federated multimodal fusion method 2D-FedSMKL studied in the main manuscript.
\medskip
\setcounter{section}{0}\setcounter{figure}{0}\setcounter{table}{0}

\noindent\textit{Notation.} $S$ denotes statistical sites and $A$ computational row groups; these need not coincide. Section~3 of the main text defines the feature, parameter and correction notation used in the proofs.
\smallskip
\noindent\textit{Terminology note.} The hybrid federated setting uses the two-dimensional sample--feature layout defined in the main manuscript. Throughout the figures in this Supplementary Material, 2D-Fed is the compact legend label for the proposed 2D-FedSMKL method.

\renewcommand{\thesection}{S\arabic{section}}
\renewcommand{\thefigure}{S\arabic{figure}}
\renewcommand{\thetable}{S\arabic{table}}

\section{Additional simulation diagnostics}

\subsection{Data-generating process and evaluation population}

The simulations use the following fixed design, implemented in
\path{code/simulate_experiments.py}. For site $s\in\{1,2,3\}$, draw
$z\sim N(h\mu_s,I_5)$, where the rows of the mean matrix are
\[
\begin{pmatrix}
0&0&0&0&0\\
.35&-.20&.15&0&.10\\
-.30&.30&-.10&.20&-.15
\end{pmatrix}.
\]
Set $\theta_0=(1,-.85,.70,.50,-.40)^\top$, and let the rows of
$D$ be $(0,0,0,0,0)$, $(-.35,.45,.10,-.20,.15)$ and
$(-1,1.10,-.90,.40,.10)$. The label is
\[
y=2\ind\{z^\top(\theta_0+hD_s)+\xi>c_s\}-1,
\qquad c=(.05,-.10,.15),
\]
with independent $\xi\sim N(0,\sigma_y^2)$. The archived design uses
$\sigma_y=1.25$ in training and $1.20$ in testing. This small train--test
noise change is part of the predictive stress design; the experiment is not
a direct verification of an i.i.d. population-risk theorem.

For modality dimensions $d=(8,12,16,12)$, generate fixed matrices
$A_m\in\mathbb R^{5\times d_m}$ with independent $N(0,1/5)$ entries
using NumPy seed 20260808. Measurements are
\[
x_m=\{1+.08h(s-2)\}z^\top A_m+h u_{sm}+e_m,
\]
where $u_{sm}$ has independent $N(0,.22^2)$ entries and is fixed using
seed $9100+100(s-1)+(m-1)$; $e_m$ has independent
$N\{0,(.60+.10m)^2\}$ entries. Training site sizes are $(360,140,60)$
and test sizes are $(300,300,300)$. For replicate seed $b$, training
and test site draws use $b+37(s-1)$ and $b+10000+53(s-1)$, respectively.

Modality 1 is always observed. Standardize its coordinates using training
means and standard deviations and use its first three coordinates in the
observation model. For $m=2,3,4$, set
\[
\begin{split}
u_{sm}&=\operatorname{logit}(1-r)+o_m+\gamma_{sm}
 +.35x_1^{(1)}-.25x_2^{(1)}+.15x_3^{(1)},\\
p_{sm}&=\operatorname{clip}[\sigma(u_{sm}),.08,.98].
\end{split}
\]
where $r$ is the target absence setting, $(o_2,o_3,o_4)=(.35,0,-.35)$,
and the rows of $(\gamma_{sm})_{m=2,3,4;s=1,2,3}$ are
$(0,0,0)$, $(0,-.15,-.25)$ and $(0,.25,-.10)$.
The target setting is not an exactly calibrated realized absence fraction.
Bernoulli masks use seed $b+8888$. A ridge-logistic observation model with
intercept, the same three baseline coordinates and two site indicators is
fitted to the training masks; its ridge is $10^{-3}$ and its intercept is
unpenalized. Estimated propensities are clipped to $[.10,.98]$.
The true-propensity lower bound is .08, not the estimated clipping threshold .10.

Each modality is standardized using its observed training rows. Zero-fill
and IPW variants multiply these coordinates by $r_m$ and
$r_m/\widehat p_m$, respectively; because the coordinates are centered, the
zero-fill variant is mean imputation. Complete-case SMKL trains on rows with
all modalities observed, retains the same SMKL penalty, and predicts all
masked test rows using zero-filled inputs. Thus its contrast combines
training-row selection and inference adaptation. Fits with fewer than
30 training rows or fewer than two observed classes are omitted; all
14 omitted fits occur for complete-case training at 65\% target absence.
All methods are evaluated on the same 900 test rows in a replicate.
All 786 successful replicate-level records are retained in the project
repository, while this Supplementary Material reports their
summaries. The Gaussian measurement model has no deterministic population
feature bound, so these predictive experiments do not establish applicability
of the bounded-feature theorem.

\subsection{Reversed and size-balanced designs}
The design-sensitivity experiments reuse the seeds and hyperparameters of the
primary simulation: seeds 16000--16019 at $h=0.8$ and 17000--17019 at $h=1.2$,
35\% absence, 20 replicates. In the reversed design the decision-boundary
offsets of the first and third sites are exchanged, so the largest site is the
hardest; in the balanced design the training sizes are $(187,187,186)$ with the
original shift assignment. Table~\ref{tab:supp_simdesign} reports the
comparison.

\begin{table}[H]
\centering
\caption{Design sensitivity of the simulation. Worst-site AUC at 35\% absence (20 replicates); the last column is the paired difference of 2D-FedSMKL against IPW mean. \emph{Original} assigns the largest decision-boundary shift to the smallest site; \emph{reversed} assigns it to the largest site; \emph{balanced} uses site sizes $(187,187,186)$.}
\label{tab:supp_simdesign}
\small
\setlength{\tabcolsep}{4pt}
\begin{tabular}{llcccc}
\toprule
Design & $h$ & 2D-FedSMKL & IPW mean & Zero-fill & $\Delta$ vs IPW mean\\
\midrule
Original & 0.8 & 0.622 & 0.612 & 0.613 & +0.010 [+0.003, +0.017] \\
Original & 1.2 & 0.535 & 0.486 & 0.483 & +0.049 [+0.038, +0.060] \\
Reversed & 0.8 & 0.669 & 0.670 & 0.674 & -0.000 [-0.004, +0.004] \\
Reversed & 1.2 & 0.604 & 0.586 & 0.588 & +0.018 [+0.009, +0.028] \\
Balanced & 0.8 & 0.659 & 0.646 & 0.651 & +0.014 [+0.005, +0.022] \\
Balanced & 1.2 & 0.606 & 0.581 & 0.585 & +0.025 [+0.014, +0.035] \\
\bottomrule
\end{tabular}
\end{table}

\begin{figure}[H]
\centering
\begin{minipage}[t]{0.49\textwidth}
\centering
\begin{tikzpicture}
\begin{axis}[width=0.92\linewidth,height=52mm,xlabel={Heterogeneity $h$},ylabel={Worst-site AUC},xmin=-0.05,xmax=1.25,ymin=0.43,ymax=0.86,grid=major,legend to name=sensitivitylegend,legend columns=4,legend style={font=\scriptsize,draw=none,/tikz/every even column/.append style={column sep=5pt}},tick label style={font=\scriptsize},label style={font=\small}]
\addplot+[mark=o] table[x=heterogeneity,y={2D-FedSMKL_mean},col sep=comma] {generated/plot_heterogeneity.csv};\addlegendentry{2D-Fed}
\addplot+[mark=square*] table[x=heterogeneity,y={IPW mean SMKL_mean},col sep=comma] {generated/plot_heterogeneity.csv};\addlegendentry{IPW mean}
\addplot+[mark=triangle*] table[x=heterogeneity,y={Zero-fill SMKL_mean},col sep=comma] {generated/plot_heterogeneity.csv};\addlegendentry{Zero fill}
\addplot+[mark=diamond*] table[x=heterogeneity,y={Complete-case SMKL_mean},col sep=comma] {generated/plot_heterogeneity.csv};\addlegendentry{Complete case}
\end{axis}\end{tikzpicture}
\end{minipage}\hfill
\begin{minipage}[t]{0.49\textwidth}
\centering
\begin{tikzpicture}
\begin{axis}[width=0.92\linewidth,height=52mm,xlabel={Target optional-modality absence},ylabel={Worst-site AUC},xmin=0.07,xmax=0.68,ymin=0.53,ymax=0.72,grid=major,unbounded coords=jump,tick label style={font=\scriptsize},label style={font=\small}]
\addplot+[mark=o] table[x=missing_rate,y={2D-FedSMKL_mean},col sep=comma] {generated/plot_missingness.csv};
\addplot+[mark=square*] table[x=missing_rate,y={IPW mean SMKL_mean},col sep=comma] {generated/plot_missingness.csv};
\addplot+[mark=triangle*] table[x=missing_rate,y={Zero-fill SMKL_mean},col sep=comma] {generated/plot_missingness.csv};
\addplot+[mark=diamond*] table[x=missing_rate,y={Complete-case SMKL_mean},col sep=comma] {generated/plot_missingness.csv};
\end{axis}\end{tikzpicture}
\end{minipage}
\par\vspace{0.4ex}\pgfplotslegendfromname{sensitivitylegend}
\caption{Simulation stress tests. The legend label 2D-Fed denotes the proposed 2D-FedSMKL method. Left: site heterogeneity at 35\% optional-modality absence. Right: controlled optional-modality absence at $h=0.8$. Points are means of 20 replicates.}
\label{fig:supp_simstress}
\end{figure}
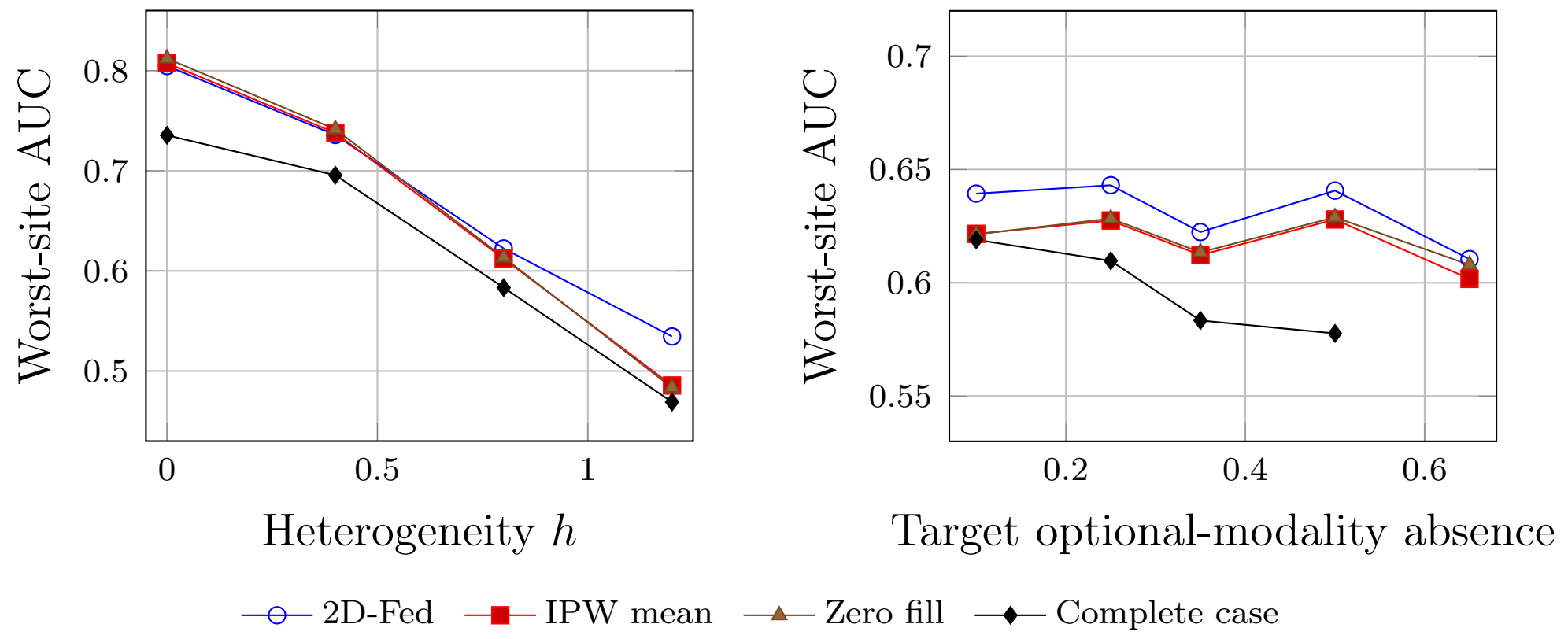

\begin{table}[H]
\centering
\caption{Paired improvement in worst-site AUC of 2D-FedSMKL over non-oracle baselines. Intervals are normal-approximation 95\% confidence intervals over 20 matched replicates.}
\label{tab:paired}
\small
\begin{tabular}{cclcc}
\toprule
Missing & $h$ & Comparator & AUC difference (95\% CI) & Win rate \\
\midrule
0.35 & 0.8 & Complete-case SMKL & 0.039 [0.023, 0.055] & 85\% \\
0.35 & 0.8 & Zero-fill SMKL & 0.009 [0.002, 0.016] & 75\% \\
0.35 & 0.8 & IPW mean SMKL & 0.010 [0.004, 0.016] & 80\% \\
0.35 & 1.2 & Complete-case SMKL & 0.066 [0.045, 0.086] & 85\% \\
0.35 & 1.2 & Zero-fill SMKL & 0.051 [0.041, 0.061] & 100\% \\
0.35 & 1.2 & IPW mean SMKL & 0.049 [0.039, 0.059] & 100\% \\
\bottomrule
\end{tabular}
\end{table}

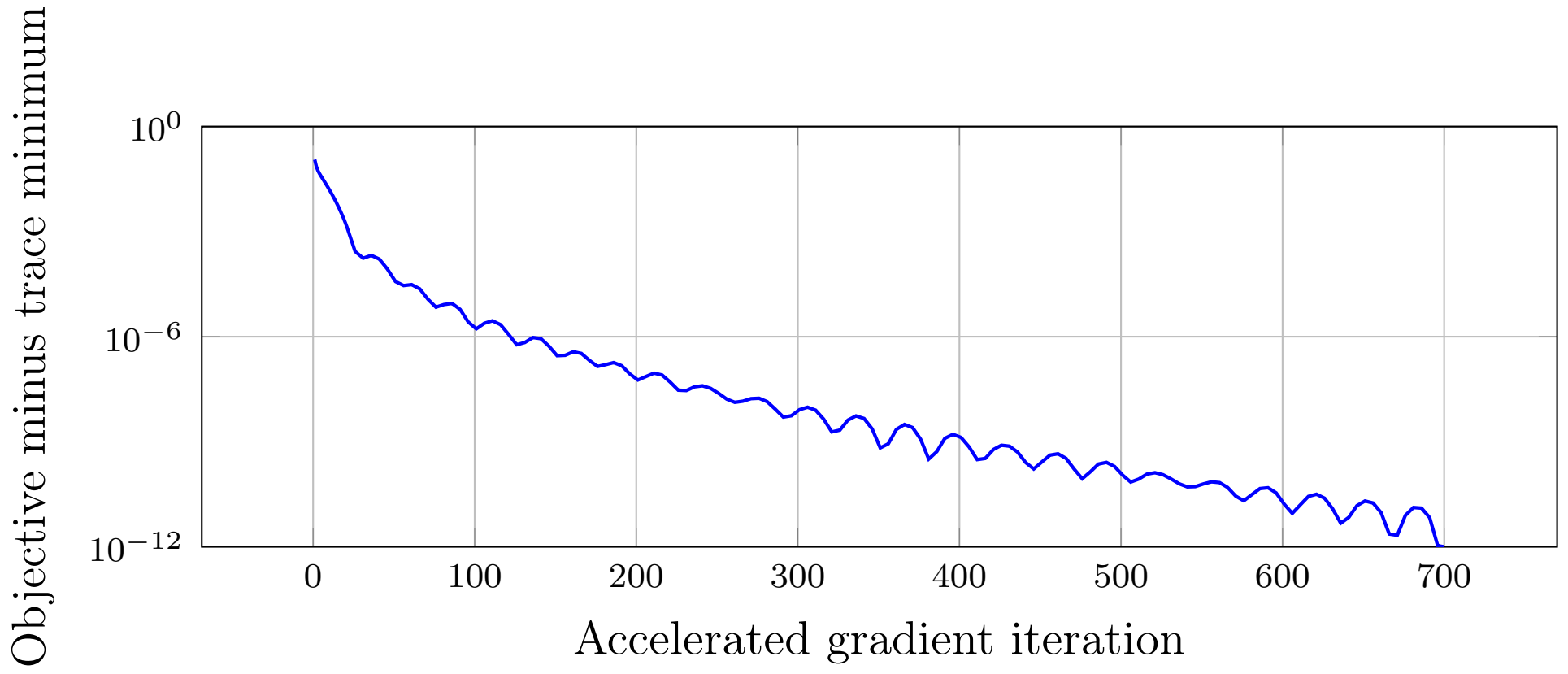
\begin{figure}[H]
\centering
\begin{tikzpicture}
\begin{semilogyaxis}[width=0.92\linewidth,height=50mm,xlabel={Accelerated gradient iteration},ylabel={Objective minus trace minimum},ymin=1e-12,ymax=1,grid=major,tick label style={font=\scriptsize},label style={font=\small}]
\addplot+[thick,no marks] table[x=iteration,y=objective_gap,col sep=comma] {figure_data/convergence.csv};
\end{semilogyaxis}
\end{tikzpicture}
\caption{Objective value minus the minimum observed over the complete reference trace, floored at $10^{-12}$ for the logarithmic axis. The reference is the best observed value over the trace, not a known $J^*$, and the plot is a numerical diagnostic.}
\label{fig:supp_convergence}
\end{figure}

\begin{table}[H]
\centering
\caption{Numerical equivalence and estimated synchronous communication for two-dimensional partitions.}
\label{tab:partition}
\scriptsize
\begin{tabular}{rrrr}
\toprule
Rows & Columns & Max. gradient error & KiB/round \\
\midrule
1 & 1 & 4.34e-17 & 0.75 \\
3 & 2 & 4.16e-17 & 11.00 \\
3 & 4 & 4.51e-17 & 28.50 \\
6 & 4 & 5.72e-17 & 30.75 \\
9 & 8 & 1.04e-16 & 68.00 \\
\bottomrule
\end{tabular}
\end{table}

\section{Public-data tuning and additional results}
\subsection{Representation fitting and source-only selection}
All public-data experiments use the same source-training representations. MHEALTH has 2335 training and 1020 test windows, four blocks, and 55 retained coordinates. For each participant/activity, the earliest 70\% of nonoverlapping 100-sample windows form the source-training set and the remainder form the test set. Thus participants are shared across partitions, and temporal dependence between adjacent windows remains possible. Within that source-training set, an earlier 80\%/later 20\% split within participant/activity is used for validation. UCI HAR retains the official 7352/2947-window subject-disjoint split, with two blocks and 50 retained coordinates. Its five training-subject silos are $\{1,3,5,6\}$, $\{7,8,11,14\}$, $\{15,16,17,19\}$, $\{21,22,23,25\}$, and $\{26,27,28,29,30\}$; test-subject silos are $\{2,4\}$, $\{9,10\}$, $\{12,13\}$, $\{18,20\}$, and $\{24\}$. Subjects 6, 14, 19, 25, and 30 are held out for inner validation, one per silo. The silos are constructed from the training subjects, so the split does not test generalization to unseen institutions.

For every inner split, blockwise means, standard deviations, and PCA are fitted exclusively on the inner-training subset and then applied to its validation subset. Final maps are refitted on the complete source-training partition and applied to the outer test partition. MHEALTH uses 15 summaries for chest acceleration, 10 for ECG, and 45 each for ankle and wrist sensing before reduction; blocks wider than 15 are reduced to 15 coordinates. HAR standardizes its 561 supplied features within the accelerometer/non-gyroscope and gyroscope blocks and retains 25 PCA coordinates per block. There is no additional standardization or whitening of the PCA coordinates. PCA uses randomized SVD with fixed seeds recorded in the script. Complete source modalities are available when these maps are fitted, before controlled removal. The experiment therefore evaluates decision fusion conditional on complete-source representations, rather than representation learning from naturally incomplete sources.

Three validation masks (integer seeds 20261131--20261133) are shared across candidates at 35\% target absence. Candidate grids are
\[
\rho\in\{0.03,0.05,0.08,0.12,0.20\},\qquad C_{\rm svm}\in\{0.1,1,10\}.
\]
FedAvg/FedProx learning rates are 0.01, 0.025 and 0.05 at 50 rounds, followed by selection among 50, 100, 200 and 350 rounds. Five local steps and the FedProx coefficient 0.1 are fixed. Selection maximizes average validation worst-site macro-F1; ties use mean-site macro-F1 and then the smaller numeric candidate. The remaining smooth parameters are fixed: $\tau=4$, $\lambda_0=0.012$, $\lambda_1=0.0035$, $\epsilon=10^{-4}$, and $\eta_b=0.05$. After training-objective and gradient diagnostics of a uniform 350-round pilot, all final primary smooth models and the $\rho$ grid use a uniform 1400-round cap, fixed from those diagnostics rather than from test performance; SVM and FedAvg/FedProx selections remain frozen. The original relative-step stopping rule may terminate a fit earlier, so terminal gradient norms are reported alongside each fit.

\begin{table}[H]
\centering\small
\caption{Frozen primary settings after source-only validation. Smooth hyperparameters not listed as selected are fixed as specified in S2.}
\label{tab:revision_selection}
\begin{tabular}{lrrrr}
\toprule
Dataset & $\rho$ & SVM $C$ & Fed learning rate & Fed rounds\\\midrule
MHEALTH & 0.12 & 10 & 0.05 & 350\\
UCI-HAR & 0.20 & 1 & 0.05 & 200\\
\bottomrule\end{tabular}
\end{table}

\subsection{Mask generation, correction, and evaluation population}
For optional block index $m=1,\ldots,M-1$ (the always-observed block has index zero), let $z$ be its first three baseline-block coordinates, standardized using source-training moments, and let $o_s$ be evenly spaced between $-0.35$ and $0.35$ across sites. Generate independent Bernoulli observations conditional on these covariates with
\[
p_{sm}(z)=\operatorname{logit}^{-1}\{\alpha_m+(0.35,-0.25,0.15)^\top z+(1+0.15m)o_s\}.
\]
The intercept $\alpha_m$ is numerically calibrated to the requested average source-training observation rate. Training-fitted ridge-logistic propensities use an intercept, the three baseline predictors and site indicators, with ridge coefficient $10^{-3}$ and unpenalized intercept. Fitted probabilities are clipped to $[0.10,0.98]$ for the default correction and applied to training and test scores. The generating probabilities are logistic probabilities without this clipping; empirical minima are documented, not treated as a population positivity proof. Masks and all correction models are regenerated for each seed. The primary study uses seeds 20260820--20260829. Every feasible method is scored on the same masked test rows. Complete-case CS trains on fully observed rows with $\lambda_1=0$, then zero-fills unavailable test blocks, which on standardized blocks is mean imputation; it does not restrict evaluation to complete test rows. Thus its comparison changes selection, regularization and inference handling together and does not isolate the effect of discarding training rows.

\begin{table}[H]
\centering
\caption{Paired differences in worst-site macro-F1 after fold-fitted source validation: 2D-FedSMKL minus comparator. Intervals are 95\% paired $t$ intervals over ten matched masks, conditional on the fixed benchmark samples.}\label{tab:public_paired}
\scriptsize
\begin{tabular}{llrr}
\toprule
Dataset & Comparator & Difference & 95\% interval \\
\midrule
MHEALTH & Zero-fill SMKL & +0.001 & [-0.024, +0.025] \\
MHEALTH & IPW mean SMKL & +0.007 & [-0.010, +0.024] \\
MHEALTH & 2D-FedSMKL no MKL & -0.001 & [-0.014, +0.011] \\
MHEALTH & Complete-case CS & +0.270 & [+0.225, +0.315] \\
MHEALTH & FedAvg-CS & +0.000 & [-0.017, +0.018] \\
MHEALTH & FedProx-CS & +0.000 & [-0.017, +0.018] \\
MHEALTH & Pooled linear SVM & +0.007 & [-0.023, +0.038] \\
UCI-HAR & Zero-fill SMKL & +0.010 & [+0.006, +0.014] \\
UCI-HAR & IPW mean SMKL & +0.000 & [-0.002, +0.002] \\
UCI-HAR & 2D-FedSMKL no MKL & +0.003 & [+0.000, +0.006] \\
UCI-HAR & Complete-case CS & +0.012 & [+0.006, +0.018] \\
UCI-HAR & FedAvg-CS & -0.001 & [-0.006, +0.003] \\
UCI-HAR & FedProx-CS & -0.001 & [-0.006, +0.003] \\
UCI-HAR & Pooled linear SVM & +0.015 & [+0.012, +0.017] \\
\bottomrule
\end{tabular}
\end{table}

\begin{table}[H]
\centering\scriptsize
\caption{Final primary smooth-model diagnostics over ten masks. The no-MKL fit is an additional primary ablation. Norms and certificates concern the empirical training objective; reaching a budget or a relative-step rule is not a claim of exact minimization. 2D-Fed denotes 2D-FedSMKL.}
\label{tab:revision_numerics}
\begin{tabular}{llrrr}
\toprule
Dataset & Method & Worst F1 (SD) & Gradient-norm range & Max gap bound\\\midrule
MHEALTH & Zero-fill & 0.856 (0.032) & [1.44e-05, 7.02e-05] & 4.10e-06\\
MHEALTH & IPW mean & 0.850 (0.032) & [1.43e-04, 1.61e-04] & 2.16e-05\\
MHEALTH & 2D-Fed & 0.857 (0.027) & [1.33e-04, 1.62e-04] & 2.18e-05\\
MHEALTH & No-MKL & 0.858 (0.031) & [1.93e-04, 2.17e-04] & 3.94e-05\\
UCI-HAR & Zero-fill & 0.777 (0.006) & [3.54e-04, 3.94e-04] & 1.30e-04\\
UCI-HAR & IPW mean & 0.787 (0.010) & [3.40e-04, 4.12e-04] & 1.41e-04\\
UCI-HAR & 2D-Fed & 0.787 (0.010) & [3.40e-04, 4.12e-04] & 1.42e-04\\
UCI-HAR & No-MKL & 0.784 (0.011) & [3.80e-04, 4.40e-04] & 1.61e-04\\
\bottomrule\end{tabular}
\end{table}

Raw metrics, per-site results, sample-level predictions and validation grids are in \path{results_revision}; the independent confusion-matrix audit reconstructs every primary metric from those predictions. Paired $t$ intervals describe variation across masks conditional on these benchmark samples, their split and their representations, without multiplicity adjustment; they do not extend to new sites or the population.

\subsection{Factorial component and propensity stress studies}
The separate, prespecified component study holds the representation and all common parameters fixed, including $\rho=0.05$ and a 350-round cap. It crosses mean/robust site aggregation with zero-filled/IPW scores at 10\%, 35\% and 65\% target absence, using 20 matched masks (seeds 20261101--20261120). At 35\% and 65\% it also removes the group penalty. These are controlled finite-budget component comparisons, not additional evaluations using the primary study's selected $\rho$ and 1400-round cap. The 2D-Fed legend label denotes robust IPW with the group penalty, namely 2D-FedSMKL.

\begin{table}[H]
\centering\small\setlength{\tabcolsep}{3.5pt}
\caption{Fixed-representation component comparison. Entries are mean (SD) worst-site macro-F1 over 20 matched masks; settings are frozen before evaluation.}
\label{tab:revision_components}
\begin{tabular}{lcccc}
\toprule
 & \multicolumn{2}{c}{MHEALTH} & \multicolumn{2}{c}{UCI HAR}\\
Method & 35\% & 65\% & 35\% & 65\%\\\midrule
Mean zero & 0.857 (0.027) & 0.769 (0.031) & 0.775 (0.008) & 0.742 (0.005)\\
Robust zero & 0.865 (0.021) & 0.787 (0.025) & 0.776 (0.008) & 0.747 (0.006)\\
Mean IPW & 0.847 (0.027) & 0.767 (0.039) & 0.781 (0.009) & 0.751 (0.007)\\
Robust IPW & 0.863 (0.030) & 0.787 (0.031) & 0.781 (0.008) & 0.755 (0.006)\\
Robust IPW no-MKL & 0.863 (0.029) & 0.806 (0.024) & 0.777 (0.010) & 0.753 (0.007)\\
\bottomrule\end{tabular}
\end{table}

\begin{table}[H]
\centering\scriptsize
\caption{Prespecified secondary component study with common $\rho=0.05$ and a 350-round cap. Entries are paired differences in worst-site macro-F1 over 20 matched masks with unadjusted 95\% $t$ intervals. Robust given IPW is robust minus mean risk under IPW, and analogously for the other simple effects. Interaction is (robust IPW minus mean IPW) minus (robust zero minus mean zero); MKL is full minus no-MKL under robust IPW. These settings are not selected by the nested primary validation.}
\label{tab:revision_factorial_paired}
\begin{tabular}{ll l rr}
\toprule
Dataset & Absence & Contrast & Difference & 95\% interval\\\midrule
MHEALTH & 35\% & Robust given IPW & +0.0157 & [+0.0063, +0.0251]\\
MHEALTH & 35\% & Robust given zero & +0.0078 & [-0.0075, +0.0232]\\
MHEALTH & 35\% & IPW given mean & -0.0102 & [-0.0234, +0.0031]\\
MHEALTH & 35\% & IPW given robust & -0.0022 & [-0.0110, +0.0065]\\
MHEALTH & 35\% & Interaction & +0.0079 & [-0.0060, +0.0219]\\
MHEALTH & 35\% & MKL given robust IPW & -0.0006 & [-0.0105, +0.0092]\\
MHEALTH & 65\% & Robust given IPW & +0.0196 & [+0.0039, +0.0354]\\
MHEALTH & 65\% & Robust given zero & +0.0176 & [+0.0096, +0.0255]\\
MHEALTH & 65\% & IPW given mean & -0.0016 & [-0.0186, +0.0153]\\
MHEALTH & 65\% & IPW given robust & +0.0004 & [-0.0137, +0.0145]\\
MHEALTH & 65\% & Interaction & +0.0020 & [-0.0154, +0.0195]\\
MHEALTH & 65\% & MKL given robust IPW & -0.0191 & [-0.0289, -0.0093]\\
UCI-HAR & 35\% & Robust given IPW & +0.0006 & [-0.0016, +0.0028]\\
UCI-HAR & 35\% & Robust given zero & +0.0007 & [-0.0013, +0.0028]\\
UCI-HAR & 35\% & IPW given mean & +0.0058 & [+0.0039, +0.0077]\\
UCI-HAR & 35\% & IPW given robust & +0.0056 & [+0.0039, +0.0074]\\
UCI-HAR & 35\% & Interaction & -0.0001 & [-0.0026, +0.0023]\\
UCI-HAR & 35\% & MKL given robust IPW & +0.0038 & [+0.0012, +0.0065]\\
UCI-HAR & 65\% & Robust given IPW & +0.0038 & [+0.0013, +0.0062]\\
UCI-HAR & 65\% & Robust given zero & +0.0055 & [+0.0024, +0.0086]\\
UCI-HAR & 65\% & IPW given mean & +0.0097 & [+0.0073, +0.0122]\\
UCI-HAR & 65\% & IPW given robust & +0.0080 & [+0.0053, +0.0107]\\
UCI-HAR & 65\% & Interaction & -0.0017 & [-0.0043, +0.0008]\\
UCI-HAR & 65\% & MKL given robust IPW & +0.0019 & [-0.0003, +0.0041]\\
\bottomrule\end{tabular}
\end{table}

\begin{table}[H]
\centering\scriptsize
\caption{Descriptive coefficient-norm shares at 65\% absence in the secondary study: mean (SD) over 20 masks. Shares use smoothed modality Frobenius norms normalized to sum to one in both fits; no-MKL shares are post-fit summaries rather than optimized MKL weights. These are not causal importance measures.}
\label{tab:revision_weight_stability}
\begin{tabular}{llrrrr}
\toprule
Dataset & Penalty & Block 1 & Block 2 & Block 3 & Block 4\\\midrule
MHEALTH & Full MKL & 0.625 (0.009) & 0.058 (0.008) & 0.171 (0.009) & 0.145 (0.007)\\
MHEALTH & No-MKL & 0.554 (0.007) & 0.103 (0.006) & 0.186 (0.007) & 0.156 (0.007)\\
UCI-HAR & Full MKL & 0.789 (0.010) & 0.211 (0.010) & -- & --\\
UCI-HAR & No-MKL & 0.761 (0.010) & 0.239 (0.010) & -- & --\\
\bottomrule\end{tabular}
\end{table}

\begin{table}[H]
\centering\scriptsize
\caption{Terminal diagnostics for all 620 prespecified secondary fits, including sensitivity variants. Gradient norm and empirical objective-gap certificate are shown as median / maximum. For this strongly convex objective, $J(\theta)-J^*\leq\|\nabla J(\theta)\|^2/(2\mu)$ with $\mu=\lambda_0\eta_b=0.0006$. The certificate concerns numerical optimization, not generalization; cap attainment is not itself convergence.}
\label{tab:revision_terminal}
\begin{tabular}{llrrrr}
\toprule
Dataset & Absence & Fits & At cap & Gradient norm & Gap certificate\\\midrule
MHEALTH & 10\% & 80 & 80/80 & 0.00154 / 0.00177 & 0.00198 / 0.00261\\
MHEALTH & 35\% & 100 & 100/100 & 0.00201 / 0.00253 & 0.00335 / 0.00532\\
MHEALTH & 65\% & 130 & 130/130 & 0.00278 / 0.00573 & 0.00642 / 0.0274\\
UCI-HAR & 10\% & 80 & 80/80 & 0.00397 / 0.00422 & 0.0131 / 0.0148\\
UCI-HAR & 35\% & 100 & 100/100 & 0.00395 / 0.00437 & 0.013 / 0.0159\\
UCI-HAR & 65\% & 130 & 130/130 & 0.00689 / 0.00792 & 0.0396 / 0.0523\\
\bottomrule\end{tabular}
\end{table}

The interaction contrast is $(\text{robust IPW}-\text{mean IPW})-(\text{robust zero}-\text{mean zero})$. It separates the two factors descriptively, rather than establishing an additive mechanism or a causal modality effect. Smoothed coefficient-norm weights sum to one and describe fitted blocks; rescaling or removing the penalty changes their interpretation. A penalty that stabilizes a coefficient pattern need not improve prediction.

At 65\% absence, the first ten matched masks are reused for propensity sensitivity: clipping floors 0.05 and 0.20 replace 0.10, or the propensity model omits the baseline covariates and uses sites only. The mask realization stays unchanged. ESS is $(\sum_i r_i/\widehat p_i)^2/\sum_i(r_i/\widehat p_i)^2$, computed within each optional modality and divided by its observed training count. Reported averages are over modality-specific ESS ratios, rather than a pooled-record ESS.

\begin{table}[H]
\centering\scriptsize
\caption{Propensity sensitivity at 65\% absence using 10 matched masks. ESS ratio is the average modality-specific effective sample size divided by observed training records. Differences compare each setting with baseline-plus-site propensity and clipping floor 0.10.}
\label{tab:revision_propensity}
\begin{tabular}{llrrrr}
\toprule
Dataset & Setting & Worst F1 & Difference & Lower clipped & ESS ratio\\\midrule
MHEALTH & Full, 0.10 & 0.798 & +0.000 & 0.0\% & 0.885\\
MHEALTH & Full, 0.05 & 0.798 & +0.000 & 0.0\% & 0.885\\
MHEALTH & Full, 0.20 & 0.799 & +0.001 & 4.7\% & 0.889\\
MHEALTH & Site-only, 0.10 & 0.796 & -0.001 & 0.0\% & 0.963\\
UCI-HAR & Full, 0.10 & 0.754 & +0.000 & 0.0\% & 0.887\\
UCI-HAR & Full, 0.05 & 0.754 & +0.000 & 0.0\% & 0.887\\
UCI-HAR & Full, 0.20 & 0.754 & +0.000 & 8.7\% & 0.908\\
UCI-HAR & Site-only, 0.10 & 0.754 & -0.000 & 0.0\% & 0.972\\
\bottomrule\end{tabular}
\end{table}

Full quantiles, true generating-propensity minima, estimated-propensity quantiles, clipping fractions, RMSE and ESS are stored in \path{component_propensity_diagnostics.csv} and its summaries. The diagnostics also record the maximum deviation $e_p$ and the resulting perturbation constant; Table~\ref{tab:supp_propdev} reports both. The sensitivity experiment is a local check under controlled MAR, not evidence of robustness to arbitrary misspecification or MNAR.

\begin{table}[H]
\centering
\caption{Observed propensity deviation and the perturbation constant of Corollary~2. $e_p=\max|\hat p-p|$ over training rows and optional modalities; $\Delta_p=B_w\kappa e_p/p_{\min}^2$ with $B_w$ from Corollary~1 and $p_{\min}=0.08$.}
\label{tab:supp_propdev}
\small
\begin{tabular}{lccccc}
\toprule
Dataset & $e_p$ min & $e_p$ max & $\Delta_p$ min & $\Delta_p$ max & $B_w$\\
\midrule
MHEALTH & 0.0799 & 0.1617 & 11451.2 & 23164.9 & 25.88 \\
UCI-HAR & 0.0208 & 0.1460 & 8580.9 & 60116.9 & 18.30 \\
\bottomrule
\end{tabular}
\end{table}

\subsection{Numerical budget sensitivity}
\begin{table}[H]
\centering
\caption{Numerical budget sensitivity on three prespecified matched masks. Changes are 1400-budget minus 350-budget values; gradients are terminal Euclidean norms. Diagnostic test metrics were not used to select the primary budget. This pilot diagnostic uses $\rho=0.12$ for MHEALTH and $0.05$ for HAR; the final primary HAR $\rho$ was subsequently reselected as $0.20$. 2D-Fed denotes 2D-FedSMKL.}\label{tab:revision_cap}
\scriptsize
\begin{tabular}{llrrrrr}
\toprule
Dataset & Method & $\Delta J$ & $\|g_{350}\|$ & $\|g_{1400}\|$ & $\Delta$ F1 & Max $|\Delta$ F1$|$ \\
\midrule
MHEALTH & 2D-Fed & -1.12e-04 & 2.04e-03 & 1.53e-04 & +0.0000 & 0.0000 \\
MHEALTH & IPW mean & -1.08e-04 & 1.99e-03 & 1.55e-04 & +0.0000 & 0.0000 \\
MHEALTH & Zero-fill & -2.13e-05 & 9.09e-04 & 5.67e-05 & +0.0031 & 0.0093 \\
UCI-HAR & 2D-Fed & -4.69e-04 & 4.04e-03 & 3.48e-04 & +0.0014 & 0.0056 \\
UCI-HAR & IPW mean & -4.12e-04 & 3.84e-03 & 3.67e-04 & +0.0025 & 0.0036 \\
UCI-HAR & Zero-fill & -4.31e-04 & 3.98e-03 & 3.82e-04 & +0.0010 & 0.0021 \\
\bottomrule
\end{tabular}
\end{table}

The diagnostic uses three fixed primary masks and the pilot temperatures: $\rho=0.12$ for MHEALTH and $0.05$ for HAR; the primary HAR temperature is $0.20$. Small test-metric differences can change with numerical accuracy and with the round budget. The final primary protocol uses the uniformly increased budget described above, while the factorial and message-interface experiments retain their common 350-round budgets. For positive $\mu=\lambda_0\min(1,\eta_b)=0.0006$, Supplementary Material~S4.8 gives the empirical objective-gap upper bound $\|\nabla J(\theta)\|_2^2/(2\mu)$. This can be conservative and is separate from predictive uncertainty and the binary population-risk bound.

\subsection{Correction comparison and validation stability}
\label{sec:supp_correction}
The conditional-imputation instantiation fits, for each optional block, a ridge regression ($\lambda=10^{-2}$, unpenalized intercept) from the always-observed block on training rows where the block is observed, and applies it to rows where it is absent; the decision layer and hyperparameters match the primary protocol. Table~\ref{tab:supp_impute} reports the 350-round comparison, Table~\ref{tab:supp_corrval} the selection between the two instantiations on ten inner-validation masks at 1400 rounds, and Table~\ref{tab:supp_rhoval} the $\rho$ grid re-run at the primary 1400-round budget on the same masks, with the selection rule of Section~S2.1. A modality-dropout training baseline removes each optional block at random during training (keep probability 0.7, three augmented copies concatenated with the original training set) and mean-fills absent blocks at test time; at 350 rounds and 20 matched masks it reaches 0.861 on MHEALTH and 0.772 on UCI-HAR, below the selected model at the same budget.

\begin{table}[H]
\centering
\caption{Correction and aggregation in the secondary budget (350 rounds, 20 masks). Worst-site macro-F1, mean over masks. At 35\% absence the conditional-imputation variant is $+0.035$ $[+0.021,+0.049]$ above IPW mean on MHEALTH and $-0.000$ $[-0.005,+0.004]$ on UCI-HAR.}
\label{tab:supp_impute}
\scriptsize
\setlength{\tabcolsep}{6pt}
\begin{tabular}{lcc}
\toprule
Method & 35\% absence & 65\% absence\\
\midrule
\multicolumn{3}{l}{\textbf{MHEALTH}}\\
Zero-fill (mean imputation) & 0.858 & 0.763 \\
Conditional imputation & 0.875 & 0.776 \\
IPW mean & 0.848 & 0.778 \\
2D-FedSMKL (IPW, robust) & 0.856 & 0.785 \\
Conditional imputation + robust & 0.884 & 0.816 \\
\midrule
\multicolumn{3}{l}{\textbf{UCI-HAR}}\\
Zero-fill (mean imputation) & 0.778 & 0.744 \\
Conditional imputation & 0.783 & 0.750 \\
IPW mean & 0.784 & 0.752 \\
2D-FedSMKL (IPW, robust) & 0.784 & 0.752 \\
Conditional imputation + robust & 0.784 & 0.748 \\
\bottomrule
\end{tabular}
\end{table}

\begin{table}[H]
\centering
\caption{Correction selection on ten inner-validation masks (1400 rounds, 35\% absence). Entries are mean worst-site macro-F1; the selection rule prefers the higher value, with ties to the IPW form.}
\label{tab:supp_corrval}
\small
\begin{tabular}{lccc}
\toprule
Dataset & IPW & Conditional imputation & Selected\\
\midrule
MHEALTH & 0.8168 & 0.8282 & Conditional imputation \\
UCI-HAR & 0.5549 & 0.5769 & Conditional imputation \\
\bottomrule
\end{tabular}
\end{table}

\begin{table}[H]
\centering
\caption{Validation of $\rho$ on ten validation masks at the full 1400-round budget, 35\% absence. Entries are mean worst-site macro-F1 over the ten masks.}
\label{tab:supp_rhoval}
\small
\begin{tabular}{lccccc}
\toprule
Dataset & $\rho=0.03$ & $0.05$ & $0.08$ & $0.12$ & $0.20$\\
\midrule
MHEALTH & 0.8016 & 0.8093 & 0.8169 & 0.8168 & 0.8089 \\
UCI-HAR & 0.5463 & 0.5481 & 0.5512 & 0.5533 & 0.5549 \\
\bottomrule
\end{tabular}
\end{table}

\section{Communication and computation audit}
The multiclass audit evaluates the actual score/gradient message interface on the shared public-data caches, at 35\% absence and mask seed 20261101. Both FISTA and plain gradient descent (GD) start at zero, use the same fixed objective ($\rho=0.05$ and the common smooth parameters in S2), initial line-search estimate 1, doubling backtracking, and 350 accepted rounds. Each candidate is recomputed after a rejected majorant. Computational row groups $A$ are independent of the $S$ statistical sites; rows may contain observations from several sites. Each of the $AB$ feature workers holds only a copy of its own matrix block. Separate row coordinators hold labels and site identifiers, and a separate global coordinator receives summaries. No central feature-matrix multiplication is used by the message evaluator. A central evaluator is used only for independent checking.

The central numerical reference is the best objective observed during an independent L-BFGS-B run (up to 4000 iterations), not a known exact $J^*$. Its terminal gradient norms are $6.26\times10^{-8}$ for MHEALTH and $7.09\times10^{-8}$ for HAR, giving empirical gap upper bounds below $4.2\times10^{-12}$. The target $10^{-3}$ is the absolute difference from that reference. Full accepted-round traces, independent reference traces, final parameter arrays, random-point gradient probes and the message ledger are included.

\begin{table}[t]
\centering
\caption{Multiclass hybrid message-interface audit on fixed public-data caches with 35\% imposed absence (seed 20261101). All methods use 350 accepted rounds and the same objective. Gap is relative to an independently optimized best observed central objective, not a known optimum. MiB counts float64 application payloads in the explicit separate-coordinator topology; it is not measured network traffic. A dash indicates that the $10^{-3}$ reference-gap target was not reached.}
\label{tab:revision_hybrid}
\small
\begin{tabular}{llrrrr}
\toprule
Data & Layout/method & Gap & $\|\nabla J\|$ & Target round & MiB \\
\midrule
MHEALTH & 1x1 FISTA & 1.12e-04 & 2.04e-03 & 188 & 232.8 \\
MHEALTH & 1x1 GD & 4.40e-02 & 6.84e-02 & -- & 232.8 \\
MHEALTH & 2x2 FISTA & 1.12e-04 & 2.04e-03 & 188 & 463.9 \\
MHEALTH & 2x2 GD & 4.40e-02 & 6.84e-02 & -- & 463.9 \\
MHEALTH & 4x4 FISTA & 1.12e-04 & 2.04e-03 & 188 & 926.0 \\
MHEALTH & 4x4 GD & 4.40e-02 & 6.84e-02 & -- & 926.0 \\
HAR & 1x1 FISTA & 4.44e-04 & 3.81e-03 & 204 & 359.1 \\
HAR & 1x1 GD & 5.30e-02 & 1.27e-01 & -- & 359.1 \\
HAR & 2x2 FISTA & 4.44e-04 & 3.81e-03 & 204 & 717.3 \\
HAR & 2x2 GD & 5.30e-02 & 1.27e-01 & -- & 717.3 \\
HAR & 4x4 FISTA & 4.44e-04 & 3.81e-03 & 204 & 1433.8 \\
HAR & 4x4 GD & 5.30e-02 & 1.27e-01 & -- & 1433.8 \\
\bottomrule
\end{tabular}
\end{table}

Across the $1\times1$, $2\times2$ and $4\times4$ layouts, the largest objective and gradient-component discrepancies at zero, random and reference points are $6.22\times10^{-15}$ and $8.70\times10^{-14}$. Each solver's distributed predictions agree with its own centralized counterpart on every test row, which leaves open whether finite-budget GD and FISTA agree with each other. FISTA reaches the reference-gap target in 188 rounds on MHEALTH and 204 on HAR; GD does not reach it in 350. This evidence concerns the specified objective, layouts and budget.

\subsection{Explicit application-payload ledger}
Each transmitted number is charged as eight bytes; broadcasts are charged once per recipient. A score-only objective pass sends parameter blocks and intercepts, partial scores, site-loss sums and modality squared-norm summaries. Its payload is
\[
V=8\{AdC+AC+BnC+AS+BM\}.
\]
A full objective/gradient pass adds site weights, weighted score derivatives, feature-gradient blocks, intercept-gradient sums, modality-penalty coefficients and the nonreplicated penalty-gradient blocks:
\[
G=V+8\{AS+BnC+AdC+AC+BM+dC\}.
\]
Here $n,d,C,M$ denote training count, feature dimension, class count and modality count. Norm and penalty reductions use the first row replica only, since parameters are replicated. With $T$ accepted rounds and $N_{\rm trial}$ backtracking trials, the total is $TG+N_{\rm trial}V$. The independent preflight recalculates every channel and the cumulative traces. In particular $1\times1$ still carries messages because the coordinators and worker remain separate. This explicit topology differs from the collapsed-coordinator approximation in the main manuscript, and the two totals are not comparable.

The FISTA payload to first reach the target is 125.56, 250.18 and 499.41 MiB on MHEALTH and 210.27, 420.07 and 839.67 MiB on HAR for the three layouts. At the common 350-round limit both solvers have the same payload in each layout because their total trial counts happen to match (355 and 357 for the respective datasets). Equal final payload does not imply equal objective accuracy. Initialization, static feature/label provisioning, test/audit calculations, headers, encryption, authentication, retransmission and secure-aggregation overhead are excluded. The runs are single-process, one-BLAS-thread emulations; no physical-network latency, parallel speedup, security or privacy guarantee is inferred.

\section{Complete mathematical proofs}
\label{app:proofs}

\subsection{Proof of convexity and smoothness}

The scalar softplus margin loss satisfies
\begin{equation*}
    \ell_\tau'(t)=-\sigma\{\tau(1-t)\},\qquad
    \ell_\tau''(t)=\tau\sigma\{\tau(1-t)\}
    [1-\sigma\{\tau(1-t)\}],
\end{equation*}
so it is convex, continuously differentiable, one-Lipschitz, and has second derivative at most $\tau/4$. Composition with an affine score preserves convexity. Consequently, each $\widehat R_s$ is convex and smooth, with $\norm{\nabla\widehat R_s}_2\leq B_x$ and Hessian operator norm at most $\tau B_x^2/4$.

The map $u\mapsto\rho\log\sum_s\pi_s\exp(u_s/\rho)$ is convex and nondecreasing in every coordinate, so its composition with the convex site risks is convex. Writing $g_s=\nabla\widehat R_s$, differentiation gives
\begin{equation*}
\begin{split}
    \nabla^2\softmaxrisk={}&\sum_s q_s\nabla^2\widehat R_s\\
    &+\rho^{-1}\left\{\sum_s q_sg_sg_s^\top-
    \left(\sum_s q_sg_s\right)\left(\sum_s q_sg_s\right)^\top\right\}.
\end{split}
\end{equation*}
The expression in braces is a covariance matrix. Its operator norm is no greater than $\sum_s q_s\norm{g_s}_2^2\leq B_x^2$, which gives the first two terms of the Lipschitz bound in Proposition~1 of the main manuscript.

It remains to bound the group penalty. Let
$g_m=(\norm{w_m}_2^2+\epsilon^2)^{1/2}$ and $v_m=w_m/g_m$. Then
\begin{equation*}
    \nabla g_m=v_m,\qquad
    \nabla^2g_m=\frac{I}{g_m}-\frac{w_mw_m^\top}{g_m^3}.
\end{equation*}
Thus $\norm{\nabla G_\epsilon}_2^2=\sum_m\norm{v_m}_2^2\leq M$ and
$\norm{\nabla^2G_\epsilon}_{\mathrm{op}}\leq1/\epsilon$. On $\norm{w}_2\leq R_w$,
\begin{equation*}
    G_\epsilon(w)\leq\sum_m(\norm{w_m}_2+\epsilon)
    \leq\sqrt{M}R_w+M\epsilon.
\end{equation*}
Since
$\nabla^2(G_\epsilon^2/2)=\nabla G_\epsilon\nabla G_\epsilon^\top+G_\epsilon\nabla^2G_\epsilon$,
its operator norm is bounded by
$2M+\sqrt{M}R_w/\epsilon$. The quadratic penalty has Hessian operator norm
$\lambda_0\max(1,\eta_b)$ and strong-convexity modulus
$\lambda_0\min(1,\eta_b)$. Summing these bounds proves Proposition~1 in the main manuscript. \hfill$\square$

\subsection{Proof of accelerated convergence}

On the convex region $\mathcal B_{R_w}$, Proposition~1 in the main manuscript and the descent lemma imply that every segment lying in the region satisfies
\begin{equation*}
    J(v)\leq J(u)+\ip{\nabla J(u)}{v-u}
    +\frac{\overline L_J(R_w)}{2}\norm{v-u}_2^2.
\end{equation*}
Hence the line-search test is guaranteed to pass whenever its trial estimate is at least $\overline L_J(R_w)$. Because a search starts from the preceding accepted estimate and multiplies only by $\gamma$, induction gives
\begin{equation*}
    \widehat L_k\leq
    \max\{\widehat L_{-1},\gamma\overline L_J(R_w)\}=L_{R_w}
    \qquad (0\leq k<T).
\end{equation*}
After each failure, Algorithm~1 recomputes the candidate at the increased estimate. Thus the accepted pair satisfies both $\theta_{k+1}=u_k-\nabla J(u_k)/\widehat L_k$ and the majorization test at that same $\widehat L_k$. The standard FISTA estimate-sequence argument for nondecreasing backtracking estimates then yields
\begin{equation*}
    J(\theta_T)-J(\theta^\star)
    \leq\frac{2L_{R_w}\norm{\theta_0-\theta^\star}_2^2}{(T+1)^2}
\end{equation*}
\cite{nesterov2005smooth,beck2009fast}, proving the accelerated-rate inequality in Theorem~1 of the main manuscript.

For the restart statement, strong convexity gives
$\norm{\theta_0-\theta^\star}_2^2\leq2\{J(\theta_0)-J(\theta^\star)\}/\mu$.
Applying the convex rate for $K$ rounds therefore multiplies the initial gap by at most
$4L_{R_w}/[\mu(K+1)^2]$. The stated value of $K$ makes this factor no greater than $1/2$. Resetting $t$ to one at every epoch and repeating the argument gives geometric contraction. \hfill$\square$

\subsection{Proof of block-gradient equivalence}

For every row $i$, the column sets form a disjoint coordinate partition, so bilinearity gives
\begin{equation*}
    \sum_{j=1}^{B}\ip{w_j}{\widetilde\phi_{ij}}
    =\ip{w}{\widetilde\phi_i}.
\end{equation*}
Thus the aggregated partial score equals the centralized score and produces the same loss, site-risk vector, adaptive weight $q_s$, and scalar loss derivative $d_i$. For any feature block $j$, the row sets are disjoint, hence
\begin{equation*}
    \sum_{a=1}^{A}\sum_{i\in\Ical_a}\omega_i d_i\widetilde\phi_{ij}
    =\sum_{i=1}^{n}\omega_i d_i\widetilde\phi_{ij},
\end{equation*}
where $\omega_i=q_{s(i)}/n_{s(i)}$. This is exactly the centralized risk gradient restricted to block $j$.

If modality $m$ spans several column blocks, disjointness also gives
$\norm{w_m}_2^2=\sum_j\norm{w_{m,j}}_2^2$ over its participating blocks. Exact reduction of these scalars therefore recovers every $g_{m,\epsilon}$ and their sum $G_\epsilon$, so the regularizer gradient in the main manuscript is identical to the centralized one. The scalar intercept derivative is likewise the sum of the weighted sample derivatives plus $\lambda_0\eta_b b$. Concatenating all blocks proves equality of the full gradients and objective values. Identical initialization, backtracking decisions, and deterministic updates then imply identical iterates by induction. \hfill$\square$

\subsection{Proof of the parameter-bound corollary}

At the zero parameter, all site risks equal $\ell_\tau(0)$ and
$G_\epsilon(0)=M\epsilon$, so
\[
J(0,0)=J_0=\ell_\tau(0)+\frac{\lambda_1}{2}M^2\epsilon^2.
\]
Optimality gives $J(\widehat w,\widehat b)\leq J_0$. Since every loss and penalty term is nonnegative,
\[
\frac{\lambda_0}{2}\sum_m\|\widehat w_m\|_2^2
+\frac{\lambda_0\eta_b}{2}\widehat b^2\leq J_0.
\]
Therefore $|\widehat b|\leq\sqrt{2J_0/(\lambda_0\eta_b)}$. Cauchy--Schwarz gives
\[\sum_m\|\widehat w_m\|_2\leq\sqrt{M}\left(\sum_m\|\widehat w_m\|_2^2\right)^{1/2},\]
and hence $\sum_m\|\widehat w_m\|_2\leq\sqrt{2MJ_0/\lambda_0}$. \hfill$\square$

\subsection{Proof of the worst-site risk bound}

Fix site $s$ and the true propensity map $p$. Because $p$ and the feature maps are fixed, the corrected scores form a fixed function class. Put
$\psi_{sim}=r_{sim}\phi_m(x_{sim})/p_m(z_{si},s)$. For Rademacher signs $\sigma_i$,
\begin{equation*}
\begin{split}
\widehat{\mathfrak R}_{n_s}(\mathcal F^p)
&=\frac{1}{n_s}\E_\sigma\sup_f
  \sum_{i=1}^{n_s}\sigma_i f^p(x_{si},r_{si},z_{si},s)\\
&\leq\frac{B_w}{n_s}\E_\sigma
  \max_m\left\|\sum_{i=1}^{n_s}\sigma_i\psi_{sim}\right\|_2
  +\frac{B_b}{n_s}\E_\sigma\left|\sum_{i=1}^{n_s}\sigma_i\right|\\
&\leq\frac{B_w\kappa/p_{\min}+B_b}{\sqrt{n_s}}.
\end{split}
\end{equation*}
For the last line, bound the maximum by the Euclidean sum over modality blocks, apply Jensen's inequality, and use
$\norm{\psi_{sim}}_2\leq\kappa_m/p_{\min}$; the intercept term follows in the same way.

Because $\ell_\tau$ is one-Lipschitz, the contraction inequality gives the same upper bound for the loss-class complexity. The parameter and feature bounds imply the envelope $L_{\max}$. A standard Rademacher deviation inequality \cite{bartlett2002rademacher} therefore yields, with probability at least $1-\delta/S$, simultaneously for every $f$,
\begin{equation*}
    R_s^p(f)\leq\widehat R_s^p(f)
    +\frac{2(B_w\kappa/p_{\min}+B_b)}{\sqrt{n_s}}
    +3L_{\max}\sqrt{\frac{\log(2S/\delta)}{2n_s}}.
\end{equation*}
Taking a union bound over sites and replacing $n_s$ by $n_{\min}$ gives a uniform bound on $\max_sR_s^p(f)$ in terms of $\max_s\widehat R_s^p(f)$. Finally, the smooth-maximum inequality in the main manuscript implies
\begin{equation*}
    \max_s\widehat R_s^p(f)
    \leq\softmaxrisk(\widehat R^p(f))+\rho\log(1/\pi_{\min}),
\end{equation*}
which proves the stated worst-site generalization bound. \hfill$\square$

\subsection{Proof of the propensity perturbation result}

On $\mathcal E_p$, both the true and post-clipping estimated propensities are at least $p_{\min}$, so
\begin{equation*}
    \left|\frac{1}{\widehat p_{sim}}-\frac{1}{p_{sim}}\right|
    =\frac{|\widehat p_{sim}-p_{sim}|}{\widehat p_{sim}p_{sim}}
    \leq\frac{e_p}{p_{\min}^2}.
\end{equation*}
For every admissible $f$, Cauchy--Schwarz and the group-norm constraint then give
\begin{equation*}
\begin{split}
    |f^{\widehat p}_{w,b}-f^p_{w,b}|
    &\leq\frac{e_p}{p_{\min}^2}
      \sum_m r_m\norm{w_m}_2\kappa_m\\
    &\leq\frac{B_w\kappa e_p}{p_{\min}^2}=\Delta_p.
\end{split}
\end{equation*}
The loss is one-Lipschitz, so
$|\widehat R_s^{\widehat p}(f)-\widehat R_s^p(f)|\leq\Delta_p$ for every site. $\softmaxrisk$ has gradient $q$ in the probability simplex and is therefore one-Lipschitz in the maximum norm. It follows that
$\softmaxrisk(\widehat R^p)\leq\softmaxrisk(\widehat R^{\widehat p})+\Delta_p$.
Intersecting $\mathcal E_p$ with the event in the worst-site generalization theorem and applying the union bound proves the estimated-propensity perturbation corollary; no independence between the two events is required. \hfill$\square$

\subsection{Deterministic properties of the multiclass objective}
\label{app:multiclass_properties}

Fix the feature and correction maps and write the multiclass parameter as $\Theta=(W,b)$, with $b\in\R^C$ and $\|W\|_{\mathrm F}^2=\sum_m\|W_m\|_{\mathrm F}^2$. The multiclass objective $J_{\rm mc}$ replaces the binary loss by the smooth Crammer--Singer loss and uses the penalties
\[
\frac{\lambda_0}{2}(\|W\|_{\mathrm F}^2+\eta_b\|b\|_2^2)
+\frac{\lambda_1}{2}\left(\sum_m\sqrt{\|W_m\|_{\mathrm F}^2+\epsilon^2}\right)^2.
\]
For $\lambda_0,\lambda_1,\eta_b\geq0$ and $\tau,\rho,\epsilon>0$, this objective is convex and continuously differentiable. If $\lambda_0,\eta_b>0$, it is strongly convex with modulus $\mu=\lambda_0\min(1,\eta_b)$ and has a unique minimizer. Under the corrected-feature bound $\|\widetilde\phi_i\|_2\leq B_x$, a valid gradient Lipschitz constant on $\|W\|_{\mathrm F}\leq R_W$ is
\begin{equation}
\begin{split}
\overline L_{\rm mc}(R_W)={}&\frac{\tau B_x^2}{2}+\frac{2B_x^2}{\rho}
+\lambda_0\max(1,\eta_b)\\
&+\lambda_1\left(2M+\frac{\sqrt M R_W}{\epsilon}\right).
\end{split}
\label{eq:mc_curvature}
\end{equation}
Exact synchronous reductions of class-score vectors, site risks, modality norms, coefficient-gradient matrices, and intercept-gradient vectors give the centralized multiclass gradient. Consequently, the accelerated-rate and exact-iterate arguments apply to $J_{\rm mc}$ with this curvature constant and the same regional and synchronization conditions.

To prove these statements, let $P$ be the softmax probability vector for logits $\tau f_y$ and $\tau(1+f_c)$, $c\ne y$. The loss equals the log-sum-exp of these logits divided by $\tau$, minus $f_y$. Its score derivatives are
\[
\nabla_f\ell_\tau^{\rm mc}=P-e_y,
\qquad
\nabla_f^2\ell_\tau^{\rm mc}=\tau\{\operatorname{diag}(P)-PP^\top\}.
\]
Here $e_y$ is the $y$th coordinate vector. The gradient norm is at most $\sqrt2$. For any vector $v$, the quadratic form of the covariance matrix is the variance of a random coordinate $v_c$ drawn with probabilities $P_c$. Its variance is at most $(\max_c v_c-\min_c v_c)^2/4\leq\|v\|_2^2/2$: the first inequality follows by centering at the midpoint of the range, and the second by Cauchy--Schwarz. Thus the loss Hessian has operator norm at most $\tau/2$.

The linear map from $\Theta$ to a class-score vector has operator norm at most $B_x$. Each site-risk gradient therefore has norm at most $\sqrt2 B_x$ and Hessian norm at most $\tau B_x^2/2$. The site-aggregate Hessian decomposition in S4.1 bounds its covariance contribution by $2B_x^2/\rho$. Vectorizing each $W_m$ leaves the group-penalty argument in S4.1 unchanged, proving \eqref{eq:mc_curvature} and strong convexity. The positive quadratic penalty also makes the objective coercive, ensuring existence of the unique minimizer. Finally, the disjoint row and column sums in S4.3 hold componentwise for every class; squared Frobenius norms likewise add over disjoint coordinate blocks. This proves exact gradient and iterate equivalence. These are deterministic properties of the fixed empirical objective, not population-risk guarantees for learned preprocessing or missingness models. \hfill$\square$

\subsection{Numerical stopping and optional optimality certificates}
\label{app:stopping}

The implemented relative-iterate rule can stop when successive parameters change little, but it does not bound the objective gap by itself. For either the binary or multiclass objective, when $\mu=\lambda_0\min(1,\eta_b)>0$, any parameter $\theta$ satisfies
\begin{equation}
0\leq J(\theta)-J(\theta^\star)
\leq\frac{\|\nabla J(\theta)\|_2^2}{2\mu}.
\label{eq:gradient_certificate}
\end{equation}
Strong convexity gives
$J(v)\geq J(\theta)+\langle\nabla J(\theta),v-\theta\rangle+\mu\|v-\theta\|_2^2/2$.
Minimizing the right-hand side over $v$ yields \eqref{eq:gradient_certificate}. The gradient must be evaluated at the returned parameter, not at its extrapolated point. For a distributed certificate, all gradient blocks and the intercept gradient must be included; an inexact gradient needs a separately justified error bound.

A diagnostic record can report the termination reason, relative iterate change, final objective and gradient norm, accepted line-search estimates, and rejected-trial counts. The gradient certificate requires one final full gradient evaluation and does not follow from a recorded relative-change tolerance. The minimizer bound in Corollary~1 covers the returned point only, so a regional bound must include the accelerated and rejected trial points separately. For any returned point with $J(\theta)\leq J_0+e$, $e\geq0$, nonnegativity of the risk and penalties gives the parameter bounds in S4.4 with $J_0+e$ replacing $J_0$; this conclusion requires neither exact optimality nor a convergence-rate claim.

\clearpage
\section{Technical comparisons and information access}
\begin{table}[H]
\centering
\caption{Technical positioning relative to selected foundations. The comparisons concern the stated formulations, rather than asserting that an existing method cannot be extended. Hybrid data ownership and sample-level modality absence are distinct: the present method assumes aligned records and known feature-block correspondence before optimization.}
\label{tab:related_hybrid}
\scriptsize
\setlength{\tabcolsep}{3pt}
\begin{tabular}{>{\raggedright\arraybackslash}p{2.05cm}>{\raggedright\arraybackslash}p{3.2cm}>{\raggedright\arraybackslash}p{3.45cm}>{\raggedright\arraybackslash}p{3.45cm}}
\toprule
Formulation & Data and target & Coordination and guarantee object & Relation to the present study \\
\midrule
SimpleMKL \cite{rakotomamonjy2008simplemkl} & Kernel combination for supervised prediction; a foundation for block-norm regularization. & Classifier and kernel-weight optimization; the weighted-norm/mixed-norm connection. & Supplies the block-norm foundation. The present target couples this principle to corrected modality scores and site-risk aggregation. \\
\addlinespace
FedHD \cite{gao2023fedhd} & Learning from hybrid data, including non-uniform holding patterns. & Local updates and gradient tracking for communication-efficient hybrid optimization. & Provides a hybrid optimization comparison. The present target explicitly couples propensity-corrected scores, group regularization and a specified site-risk aggregate. \\
\addlinespace
Zhang et al.\ \cite{zhang2024hybrid} & Sample/feature heterogeneity with global/local model matching. & Feature-extractor/classifier coordination through model matching. & Addresses representation/model compatibility; the present interface assumes feature correspondence and does not perform that matching. \\
\addlinespace
2D-FedSMKL & Aligned row/column partitions; fixed representations and propensity corrections; smooth worst-site decision fusion. & Partial scores, site-risk scalars, norm summaries and block gradients. Exact synchronous reductions recover the common objective and gradient. & Target-specific convexity/curvature and equivalence properties; conditional accelerated convergence and an oracle-propensity binary risk bound. Guarantees concern the stated objective and exact synchronous reductions. \\
\bottomrule
\end{tabular}
\end{table}

\begin{table}[H]
\centering
\caption{Information access and roles of the predictive references and computational controls. All feasible predictors are evaluated on the same masked test population; complete-case methods are not scored only on a complete-test subset. Centralized/horizontal references do not independently establish feasibility under column-wise feature custody.}
\label{tab:baseline_information}
\scriptsize
\setlength{\tabcolsep}{3pt}
\begin{tabular}{>{\raggedright\arraybackslash}p{2.65cm}>{\raggedright\arraybackslash}p{5.0cm}>{\raggedright\arraybackslash}p{4.5cm}}
\toprule
Family & Training information and inference handling & Comparison role \\
\midrule
Pooled linear SVM & Central access to all masked feature vectors and labels; zero filling during training and inference. & Statistical reference under centralized access. \\
\addlinespace
Complete-case CS/SMKL & Fit on rows with every optional block observed; infer on all masked test rows after zero filling. The public CS implementation sets the MKL penalty to zero. & Joint training-selection/inference strategy; its gap does not isolate loss of training rows alone. \\
\addlinespace
FedAvg-CS/FedProx-CS & Horizontal site learners have complete zero-filled feature vectors for their own rows; exchange model parameters. & Horizontal statistical references; they do not enforce separate column custodians. \\
\addlinespace
Mean/robust $\times$ zero-fill/IPW & The same fixed representation and masked rows; cross site aggregation (mean/robust) with test and training score handling (zero-fill/IPW). & Factorial controls for aggregation, availability correction and their interaction. \\
\addlinespace
No-MKL & Same corrected inputs and robust site objective as the full method, with $\lambda_1=0$. & Control for group-kernel regularization; group weights describe the fitted coefficients rather than causal modality importance. \\
\addlinespace
Oracle full SMKL & Central access to complete training and test modalities, without artificial removal. & Infeasible full-information reference, not a deployment competitor. \\
\addlinespace
Row/column FISTA and GD & Each feature worker holds its own matrix block; row coordinators hold labels/site identifiers. Both solvers use the same corrected objective, initialization and explicit message interface. & Computational controls for objective accuracy and message payload. Emulated messages are not measurements of physical-network latency. \\
\bottomrule
\end{tabular}
\end{table}

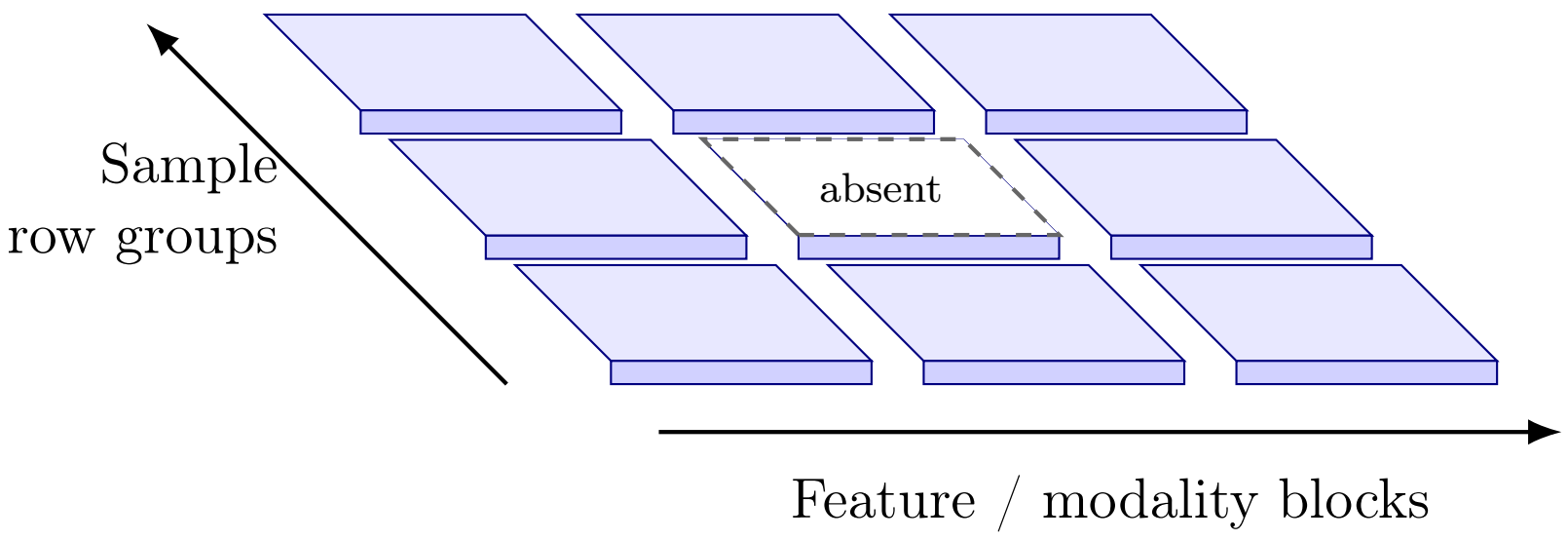
\begin{figure}[H]
\centering
\begin{tikzpicture}[x={(1.18cm,0cm)},y={(-0.5cm,0.5cm)},z={(0cm,0.45cm)},font=\small]
\foreach \i in {0,1,2}{\foreach \j in {0,1,2}{
\filldraw[fill=blue!9,draw=blue!50!black] (\i*1.8,\j*1.7,0.35) -- (\i*1.8+1.5,\j*1.7,0.35) -- (\i*1.8+1.5,\j*1.7+1.3,0.35) -- (\i*1.8,\j*1.7+1.3,0.35) -- cycle;
\filldraw[fill=blue!18,draw=blue!50!black] (\i*1.8,\j*1.7,0) -- (\i*1.8+1.5,\j*1.7,0) -- (\i*1.8+1.5,\j*1.7,0.35) -- (\i*1.8,\j*1.7,0.35) -- cycle;
}}
\filldraw[fill=white,draw=black!60,dashed,thick] (1.8,1.7,0.36) -- (3.3,1.7,0.36) -- (3.3,3,0.36) -- (1.8,3,0.36) -- cycle;
\node[font=\scriptsize] at (2.55,2.35,0.36) {absent};
\draw[-{Latex},thick] (0,-.65,0)--(5.2,-.65,0) node[midway,below=4pt] {Feature / modality blocks};
\draw[-{Latex},thick] (-.6,0,0)--(-.6,4.9,0) node[midway,left=5pt,align=right] {Sample\\row groups};
\end{tikzpicture}
\caption{Spatial schematic of the two-dimensional ownership layout. Rows partition samples and columns partition feature/modality blocks; the dashed tile illustrates an unavailable block. Tile thickness is illustrative, not a third partition dimension, sample size, risk or measured quantity. Computational row groups need not coincide with statistical sites.}
\label{fig:spatial_layout}
\end{figure}

\begingroup
\fontsize{9pt}{11pt}\selectfont

\endgroup
\end{document}